\documentclass[11pt]{article}
\usepackage{textgreek}
\usepackage{booktabs}
\usepackage{multirow}
\usepackage{stfloats}
\usepackage{tabularx}
\usepackage{subcaption}
\usepackage[table]{xcolor}
\usepackage{enumitem}
\usepackage{amsmath}
\usepackage{algorithm}
\usepackage{algpseudocode}
\usepackage{graphicx}
\usepackage[final]{acl}
\usepackage{times}
\usepackage{latexsym}

\usepackage[T1]{fontenc}

\usepackage[utf8]{inputenc}

\usepackage{microtype}

\usepackage{inconsolata}
\usepackage{svg}

\usepackage{soul}

\definecolor{myblue}{rgb}{0.651, 0.808, 0.890}
\definecolor{mygreen}{rgb}{0.698, 0.875, 0.541}
\definecolor{myorange}{rgb}{0.992, 0.749, 0.435}

\title{MultivationBench: A Benchmark for Multimodal Sequential Motivation Reasoning}

\newcommand{\ust}{\ensuremath{^\spadesuit}}
\newcommand{\amazon}{\ensuremath{^\clubsuit}}

\author{Kawai Chung\ust 
\ Chunkit Chan\ust
\ Yauwai Yim\ust
\ Yuxuan Liu\ust 
\ \ Haochen Shi\ust \\
\ \textbf{Weiqi Wang\ust} 
\ \textbf{Qing Zong\ust} 
\ \textbf{Tianshi Zheng\ust} 
\ \textbf{FU Yixuan\ust} 
\ \textbf{Wong Kai Chung\ust} \\
\ \textbf{Hao Liang\ust} 
\ \textbf{Yifan Gao\amazon}
\ \textbf{Xi Yang\ust}
\ \textbf{Janet Hui-wen Hsiao\ust}\\
\ \textbf{Yangqiu Song\ust}\\
\ust The Hong Kong University of Science and Technology\\
\amazon Amazon.com\\
\href{https://github.com/HKUST-KnowComp/MultivationBench}{\raisebox{-0.1ex}{\includegraphics[height=1.0em]{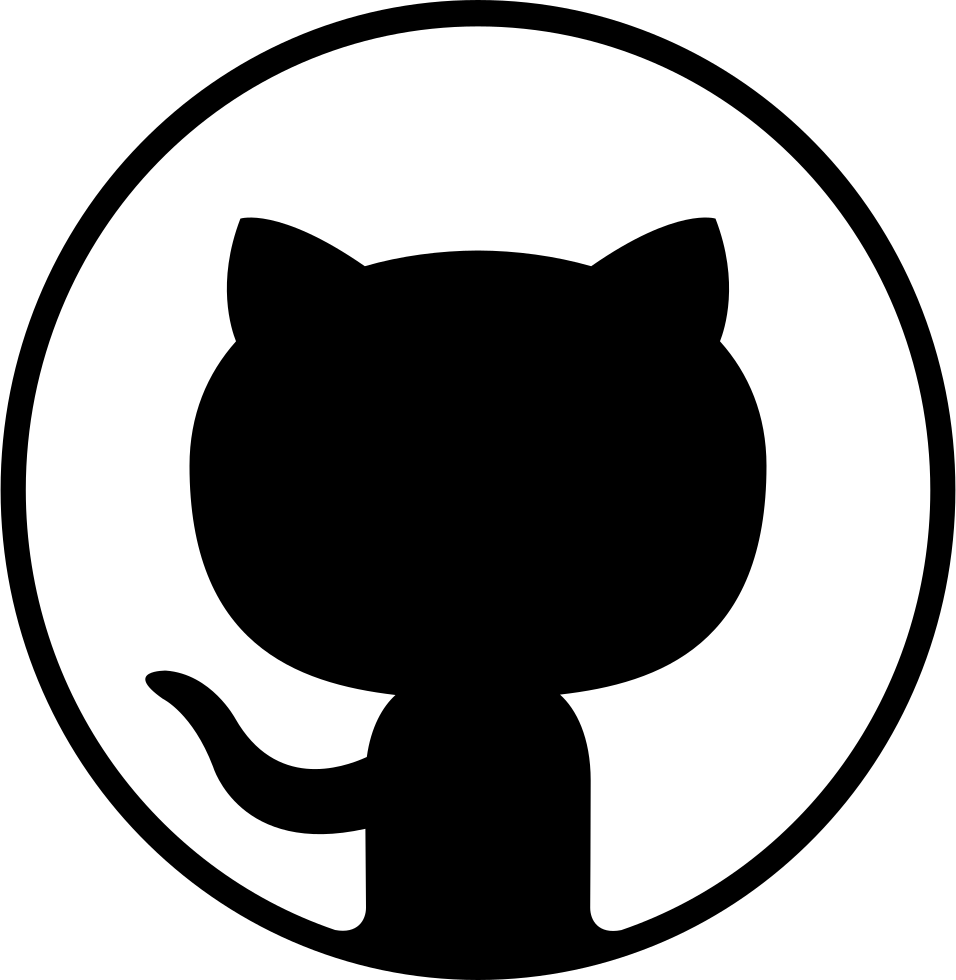}~Code}}
\texttt{\{kwchungac, yqsong\}@cse.ust.hk} \ \ \ \ 
}

\begin{document}
\maketitle
\begin{abstract}

Multimodal Large Language Models have sparked significant interest due to their potential for social intelligence; however, their ability to perform sequential motivation reasoning remains insufficiently studied. Existing evaluations predominantly examine static text or isolated visual snapshots, which do not reflect the cumulative nature of real-world behavioral drivers. To address this gap, we introduce \textsc{MulTivationBench}, a benchmark designed to rigorously evaluate multimodal motivation reasoning within story-driven visual narratives. The benchmark builds upon established psychological frameworks---Maslow's hierarchy and Reiss's basic desires---and requires models to integrate accumulated multimodal context to infer evolving motivations. Results indicate that \textsc{MulTivationBench} presents a significant challenge: all tested models struggle to maintain consistent motivation reasoning across sequential contexts, revealing a critical disconnect between static recognition capabilities and the dynamic reasoning essential for human-like social understanding.

\end{abstract}

\begin{figure*}[htb] 
    \centering
    \includegraphics[width=1\textwidth]{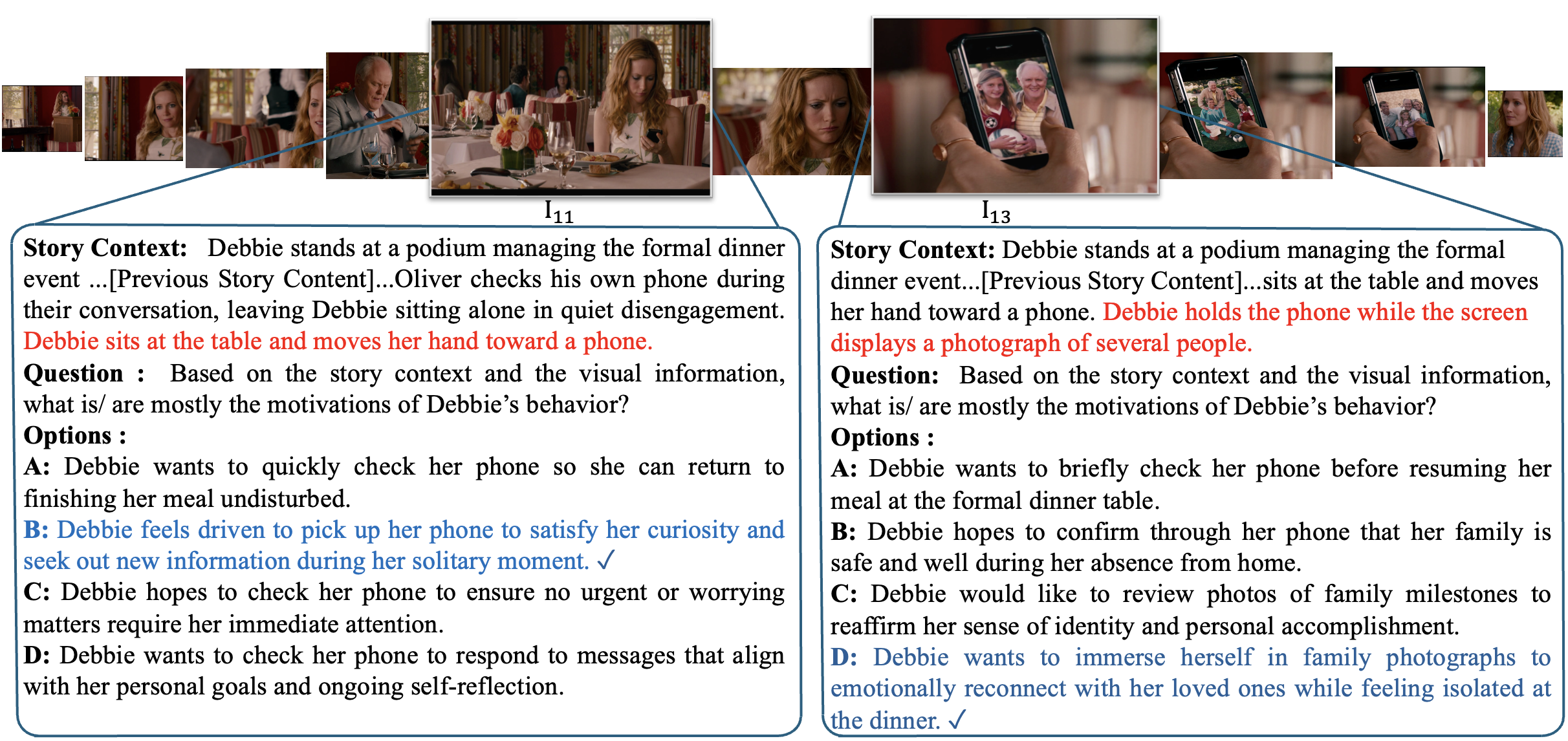}
    \caption{A Maslow motivation example from \textsc{MultivationBench}. Given the accumulated story context and images up to \(I_{11}\), Debbie reaching for her phone is plausibly interpreted as reflecting \textit{Cognitive} needs; with additional context up to \(I_{13}\), which reveals family photographs and her social isolation at the dinner, the inferred motivation shifts to \textit{Love and Belonging}. This example shows how the inferred practical motivation changes as sequential multimodal context accumulates.}
    \label{fig:multivation_overview}
\end{figure*}

\section{Introduction}
\label{sec:intro}

Motivation is the internal force that initiates, guides, and sustains goal-directed behavior, fundamental to explaining human action~\citep{hagger2005social, ryan2000self}. 
It helps explain why people begin, persist in, or stop a behavior as situations unfold~\citep{kazdin2000encyclopedia}. Yet motivation is latent, which cannot be directly observed. Instead, it must be inferred from behavior and context. For this reason, people rarely interpret behavior from a single static moment alone~\citep{scholl2000perceptual}. 
As events unfold, people accumulate evidence across time, using later actions and the surrounding context to revise earlier interpretations. 
This sequential reasoning process is critical in narratives. The motivation for an action often emerges only after additional visual and situational information is disclosed~\citep{zack2007, baldassano2017discovering, Baker2017, knoblich2008}.

However, existing motivation reasoning benchmarks focus on static text-based scenarios or isolated multimodal snapshots~\citep{socialQa, YongLY0025}. 
Such settings may reward locally plausible guesses, but do not assess whether a model can update its interpretation as a story develops. 
Therefore, current benchmarks provide limited evidence on whether Multimodal Large Language Models (MLLMs) \citep{mllms} can perform sequential motivation reasoning from an accumulated visual and textual context.

To bridge this gap, we present \textsc{MultivationBench}, a large-scale benchmark specifically designed to stress-test multimodal sequential motivation reasoning in story-driven narratives. Our benchmark is grounded in two complementary psychological frameworks: Maslow's Expanded Hierarchy of Needs~\citep{maslow1970motivation}, which captures broad categories of human needs, and Reiss's 16 Basic Desires~\citep{reiss2004}, which captures finer-grained differences in motivational content. 
This design enables evaluation at both a coarse-grained, universal level and a fine-grained, individual level, requiring models to reason over accumulated multimodal context, not just isolated observations.

Figure~\ref{fig:multivation_overview} illustrates the core challenge targeted by \textsc{MultivationBench}. 
A sequence of images shows Debbie at a formal dinner event. 
At image $I_{11}$, she reaches toward a phone. 
Viewed in isolation, this moment is ambiguous. A model might interpret the behavior as information seeking, aligning with \textit{Cognitive} needs in Maslow's hierarchy. However, later frames reveal family photograph on the phone. The broader narrative establishes Debbie's social isolation. The explanation then shifts toward seeking emotional connection with absent loved ones, which aligns with \textit{Love and Belonging}. The challenge is not simply recognizing the action, but revising the inferred motivation as new evidence accumulates. Models relying on surface-level cues may choose simpler explanations, missing motivations that require a longer multimodal context.

To benchmark this, \textsc{MultivationBench} has 1,000 unique stories, 4,023 visually grounded behavioral instances, and 16,092 evaluation questions. 
Experiments on state-of-the-art MLLMs show that sequential motivation reasoning remains challenging. Several models improve from short to medium-length stories, suggesting that moderate additional context can help in some cases. However, this benefit does not persist for longer narratives, and story-level consistency remains low across models, indicating that maintaining correct motivation reasoning across the whole narrative is still difficult.
Our contributions are summarized as follows:
\begin{itemize}[leftmargin=*]
    \item To the best of our knowledge, \textsc{MultivationBench} is the first human-annotated benchmark for evaluating multimodal sequential motivation reasoning in story-driven visual narratives.
    \item \textsc{MultivationBench} grounds motivation labels in two complementary psychological frameworks---Maslow's Expanded Hierarchy of Needs and Reiss's 16 Basic Desires---thereby enabling evaluation of both coarse-grained needs and fine-grained desires under accumulated multimodal context.
    \item We conduct systematic experiments on state-of-the-art MLLMs under full multimodal, text-only, and image-only settings, and provide in-depth analyses of local task accuracy, story-level consistency, taxonomy granularity, and the gap between model and human performance.
\end{itemize}

\section{Related Work}
\subsection{Text-Based Social and Motivation Reasoning.}
Research on social and motivational reasoning has predominantly relied on text-based evaluations. Prior work has studied event-centric intent and reaction inference~\citep{sap2019atomic,rashkin2018event2mind}, character psychology in simple stories~\citep{rashkin2018naive}, and contextualized story explanation~\citep{mostafazadeh2020glucose}.  SocialIQA~\citep{socialQa} evaluates social commonsense by presenting daily scenarios and asking models to reason about actions and their implications. While it scales well, its reliance on purely textual inputs limits its ability to assess how models interpret non-verbal social signals. More recently, MotiveBench~\citep{YongLY0025} has explicitly targeted motivation reasoning by integrating detailed character profiles and behavioral descriptions to infer psychological drives. However, it relies exclusively on textual context and typically presents isolated, single-turn scenarios. This unimodal, static approach overlooks the critical role of visual cues and fails to test whether a model can track a character's evolving motivation across a continuous narrative sequence.

\paragraph{Multimodal Social Reasoning}
Recent multimodal benchmarks have begun to incorporate visual evidence into socially grounded reasoning and temporal visual understanding~\citep{lin2025valphasocial,villacueva2025moments,jin2024mmtomqa,li2024mvbench,wu2024longvideobench,ren2024timechat,maaz2024videochatgpt}. While prior works address visual social commonsense, multimodal theory-of-mind reasoning, video understanding, temporal grounding, and visual goal inference, they are often restricted to short-horizon, event-centered, or perception-oriented settings that lack the depth to model complex, evolving psychological states. These approaches rely on surface-level behavior-to-intent mapping rather than inferring latent desires through a grounded psychological framework, thereby limiting their ability to evaluate motivation reasoning that unfolds over accumulated multimodal narrative context~\citep{chen2025}. Given the limitations in prior work, there is a need for evaluating motivation within realistic multimodal settings, capturing authentic social interactions beyond goals and beliefs alone.

Recent benchmarks cover grounded social reasoning, video interactions, nonverbal behavior, gestures, social norms, aligned interaction representations, and sequential visual prediction~\citep{mathur-etal-2025-social,kong2025sivbench,li2025mimeqa,Cao_2025_CVPR,rezaei-etal-2025-egonormia,Lee_2024_CVPR,zhu2023sequential}. \textsc{MulTivationBench} instead combines accumulated visual narratives with explicit multi-label motivation taxonomies and evaluates coherence as a story develops. Appendix~\ref{Appendix_for_Related_Works} gives a detailed comparison.

\section{\textsc{MultivationBench}}

During the construction of \textsc{MultivationBench}, we follow four core design principles: (1) \textit{Psychological grounding}, employing authoritative frameworks to define a label space beyond surface-level intents; (2) \textit{Sequential human-centric dynamics}, prioritizing narratives with rich interpersonal interactions where sequential visual cues are essential for tracking mental states; (3) \textit{Long-term consistency evaluation}, stratifying data into distinct intervals based on sequence length to evaluate a model's ability to maintain psychological continuity over different memory horizons; and (4) \textit{Data integrity}, assessing and mitigating potential data contamination risk.

\subsection{Data Sources}

Following these principles, MULTIVATIONBENCH prioritizes sequential visual narratives over static scenes, since character motivations often emerge only through later visual and situational evidence. We aggregate data from three complementary sources---\textit{SSID}~\citep{malakan2023ssid}, \textit{StoryReasoning}~\citep{oliveira2025storyreasoning}, and \textit{MovieBench}~\citep{wu2024moviebench}---to cover diverse narrative styles and temporal complexity. SSID provides shorter social-media-style stories, while StoryReasoning and MovieBench offer longer cinematic narratives with richer character arcs. We also conduct a contamination analysis and find low risk across all sources; details are in Appendix~\ref{sec:contamination}.

\subsection{Psychological Frameworks and Task Formulation}

\paragraph{Maslow's Expanded Hierarchy of Needs}
We adopt the 8-level version of Maslow's Expanded Hierarchy of Needs~\citep{maslow1970motivation} to represent universal levels of human needs that can motivate behavior. The eight labels are \textit{Physiological}, \textit{Safety}, \textit{Love and Belonging}, \textit{Esteem}, \textit{Cognitive}, \textit{Aesthetic}, \textit{Self-Actualization}, and \textit{Transcendence}. In \textsc{MultivationBench}, Maslow serves as a theory of need priority to capture whether a behavior is primarily driven by immediate survival and security concerns, by social and esteem needs, or by higher-order growth and meaning-oriented motives.

\paragraph{Reiss's Basic Desires and their relationship to Maslow}
We additionally incorporate Reiss's theory of 16 basic desires through the Reiss Motivation Profile (RMP)~\citep{reiss2004}. The sixteen labels are \textit{Physical Exercise}, \textit{Eating}, \textit{Order}, \textit{Saving}, \textit{Tranquility}, \textit{Romance}, \textit{Family}, \textit{Acceptance}, \textit{Social Contact}, \textit{Independence}, \textit{Vengeance}, \textit{Honor}, \textit{Power}, \textit{Status}, \textit{Curiosity}, and \textit{Idealism}. Unlike Maslow, which organizes motivation as universal hierarchical levels of need, Reiss emphasizes individual differences in the relative strength of specific desires that shape behavior. The two theories, therefore, serve different but complementary purposes in our benchmark: Maslow identifies the broad level of need a behavior serves, whereas Reiss distinguishes the more specific desire content that may motivate the same behavior. We use them as complementary operational taxonomies rather than exhaustive theories; alternatives such as Self-Determination Theory remain future extensions. Appendix~\ref{appendix:A} discusses the choice, alternatives, and full definitions.

\paragraph{Task formulation}
Based on these two psychological frameworks, \textsc{MultivationBench} evaluates motivation reasoning on visually grounded character behaviors in sequential visual narratives. For each visually grounded behavior exhibited by a character at a particular image index, we use the accumulated story context and images up to that point to formulate four tasks: Maslow Definition, Maslow Practical Motivation, Reiss Definition and Reiss Practical Motivation. The \textit{Definition} tasks ask models to classify the motivation of the behavior based on the standard theory labels, whereas the \textit{Practical Motivation} tasks ask models to choose among context-specific options derived from the same theories but instantiated in the current narrative situation. Some behaviors can be supported by more than one valid motivation under the story context. We, therefore, formulate all four tasks as multi-label prediction problems, where each instance may have one or more ground-truth annotations. Appendix~\ref{appendix:options} provides detailed controls for Practical-option length, lead-in style, and label-frequency imbalance; Section~\ref{sec:limitations} discusses residual risks from combined lexical cues.

\begin{figure*}[t]
    \centering
    \includegraphics[width=1\textwidth]{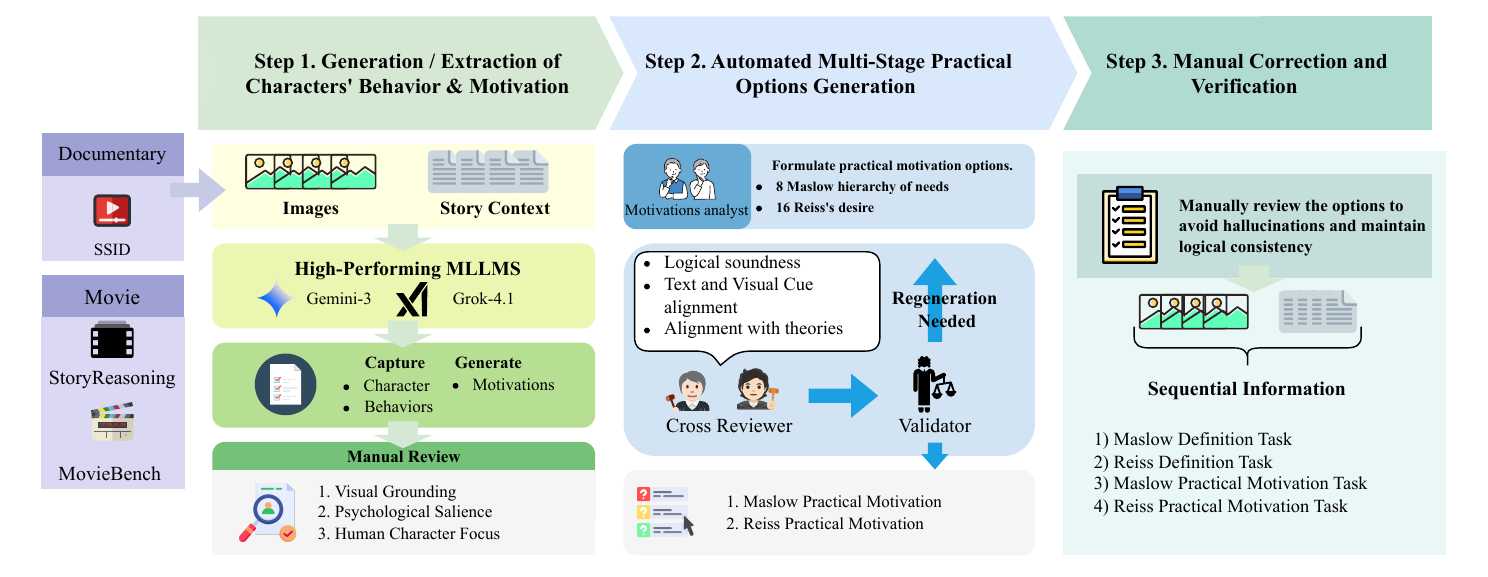}
    \caption{AI--human pipeline for constructing \textsc{MulTivationBench}. We summarize the stages here and defer prompts and implementation details in Appendix \ref{Appendix:DataConstruction}.}
    \label{fig:pipeline}
\end{figure*}

\label{sec:method}

\subsection{Benchmark Construction Overview}
\label{subsec:pipeline}

\paragraph{Data Construction}
We construct \textsc{MulTivationBench} through a compact AI--human pipeline (Figure~\ref{fig:pipeline}). For each narrative, multiple MLLMs first extract a character-centered \textit{behavior chain}: a list of visually grounded behaviors aligned to image indices with candidate motivations. Based on the verified behavior chain and the story context up to each behavior, we then generate practical motivation options under Maslow and Reiss's definitions. These candidate options are further refined through AI-based cross-review and validation to ensure logical consistency, multimodal grounding, and alignment with the psychological theories. Finally, human filtering is applied to remove hallucinated or weakly grounded samples. Across the original and replacement batches, the pipeline produced 4{,}382 behavior candidates and retained 4{,}023 after correction, filtering, and final allow-listing; 32 story slots were replaced or had their character chains corrected, and 101 question records were regenerated. The retained behaviors were checked for visual grounding, psychological salience, and consistent focus on the main human character. Detailed prompts for behavior extraction, option generation, AI review, the human filtering protocol, and the complete construction flow are described in Appendix~\ref{Appendix:DataConstruction}.

\paragraph{Human Annotation and Agreement}
We conduct a human annotation study to establish the ground-truth labels for the four reasoning tasks. Three annotators, all computer-science graduate students from English-speaking universities with prior experience in Theory-of-Mind benchmark annotation, were recruited for the annotation process. To ensure annotation quality, each annotator first completed a training phase covering the task definitions and representative examples. A researcher with a psychology background provided task-specific instruction on the Maslow and Reiss categories. We then evaluated annotator performance on the first 100 instances and provided detailed feedback on typical errors to calibrate their understanding. Fleiss's $\kappa$ values, reported on a percentage scale, are $80.99\%$, $78.87\%$, $74.87\%$, and $75.35\%$ for Maslow Definition, Maslow Practical Motivation, Reiss Definition, and Reiss Practical Motivation, respectively~\citep{fleiss1971measuring}. More details of the annotation process are provided in Appendix~\ref{appendix_annotation}.

\subsection{Dataset Statistics}
\textsc{MulTivationBench} contains 1{,}000 visual narratives with 4{,}023 visually grounded character behaviors. Each behavior is paired with four questions, yielding 16{,}092 evaluation instances. To evaluate long-horizon coherence, we stratify stories by maximum narrative length: Short ($\leq 5$ images), Medium (6--9), and Long (10+). Appendix~\ref{Appendix:stats} reports gold label-set cardinality distributions and stratified performance.

\section{Experiment}

\subsection{Experimental Setting}
We assess the motivation reasoning capabilities of eight Multimodal Large Language Models. For closed-source models, we evaluate o4-mini~\citep{openai_o4mini_2025}, Gemini-3-Flash~\citep{google2025gemini3flash}, and Grok-4.1-Fast~\citep{xai2025grok41}. For open-source models, we select the Llama-4 family (Scout-17B and Maverick-17B)~\citep{meta2025llama4}, the Phi family (Phi-3.5-Vision-instruct~\citep{abdin2024phi3technicalreporthighly} and Phi-4-multimodal-instruct~\citep{microsoft2025phi4multi}), and Nemotron-Nano-12B-v2-vl~\citep{nvidia2025nemotronnano}. All evaluations are conducted in a zero-shot setting with the temperature parameter set to $0$ to ensure reproducibility. To mitigate position bias, we randomly shuffle the order of the answer options for every instance~\citep{Zheng0M0H24}.
We also report human performance from a separate group of three computer-science graduate students who completed the same evaluation tasks and had not participated in gold-label construction. More details are provided in Appendix~\ref{App:human_evaluation}. For Phi-3.5-Vision-instruct, the Long-story multimodal and image-only results are not reported due to limited image context length, making its overall score less directly comparable. A common-subset comparability check is provided in Appendix~\ref{App:sub_common}.

\paragraph{Metrics}
We report two complementary metrics.
\textit{Exact Match (EM)} measures strict set-level accuracy: a prediction is counted as correct only when the predicted option set exactly matches the ground-truth set.
\textit{Example-based F1 (F1)} measures instance-level partial agreement in the multi-label setting.
Following prior work on multi-label evaluation, we evaluate each test instance separately and then average across instances~\citep{zhang2014review,lozamencia2023treebased}.
This metric rewards partial overlap between the predicted and ground-truth sets while penalizing both false positives and false negatives.
We provide the formal definition and an illustrative example in Appendix~\ref{app:metric-details}.

\begin{table*}[ht]
\centering
\resizebox{0.9\textwidth}{!}{\setlength{\tabcolsep}{9pt}
\begin{tabular}{l|cc|cc|cc|cc}
\toprule
\multirow{2}{*}{\textbf{Model (Modality)}} &
\multicolumn{2}{c|}{\textbf{Short ($\leq 5$)}} &
\multicolumn{2}{c|}{\textbf{Medium (6--9)}} &
\multicolumn{2}{c|}{\textbf{Long (10+)}} &
\multicolumn{2}{c}{\textbf{Overall}} \\
\cmidrule(lr){2-3} \cmidrule(lr){4-5} \cmidrule(lr){6-7} \cmidrule(lr){8-9}
 & EM(\%) & F1(\%) & EM(\%) & F1(\%) & EM(\%) & F1(\%) & EM(\%) & F1(\%) \\
\midrule
\midrule
{\footnotesize Grok-4.1-Fast}                 & 31.3 & \textbf{54.6} & 33.0 & \textbf{55.6} & 31.0 & \textbf{54.5} & 31.9 & \textbf{55.0} \\
{\footnotesize Grok-4.1-Fast (Text-Only)}     & 30.8 & \underline{53.1} & 30.9 & \underline{54.6} & 30.0 & \underline{54.1} & 30.6 & \underline{54.0} \\
{\footnotesize Grok-4.1-Fast (Image-Only)}    & 23.0 & 39.2 & 23.2 & 35.6 & 23.5 & 36.7 & 23.2 & 37.0 \\
\midrule
{\footnotesize Gemini-3-Flash}                & 31.6 & 52.8 & \underline{38.0} & 54.5 & \underline{37.0} & 53.7 & \underline{35.7} & 53.7 \\
{\footnotesize Gemini-3-Flash (Text-Only)}    & 15.1 & 50.2 & 13.7 & 49.0 & 12.0 & 48.3 & 13.6 & 49.2 \\
{\footnotesize Gemini-3-Flash (Image-Only)}   & 22.9 & 39.8 & 28.6 & 37.1 & 29.4 & 38.2 & 27.1 & 38.2 \\
\midrule
{\footnotesize o4-mini}                       & 26.9 & 52.8 & 27.5 & 53.7 & 25.1 & 52.7 & 26.6 & 53.1 \\
{\footnotesize o4-mini (Text-Only)}           & 23.7 & 52.1 & 24.2 & 52.8 & 22.1 & 51.4 & 23.5 & 52.2 \\
{\footnotesize o4-mini (Image-Only)}          & 17.6 & 39.5 & 17.1 & 35.5 & 17.0 & 35.7 & 17.2 & 36.8 \\
\midrule
{\footnotesize Llama-4-Scout-17B}             & \underline{32.5} & 48.5 & 33.8 & 50.4 & 31.3 & 47.9 & 32.7 & 49.1 \\
{\footnotesize Llama-4-Scout-17B (Text-Only)} & 28.8 & 48.1 & 26.3 & 49.1 & 25.1 & 47.8 & 26.7 & 48.4 \\
{\footnotesize Llama-4-Scout-17B (Image-Only)}& 25.4 & 33.4 & 30.4 & 34.8 & 29.2 & 34.3 & 28.5 & 34.2 \\
\midrule
{\footnotesize Llama-4-Maverick-17B}          & 26.3 & 48.1 & 32.6 & 50.5 & 31.3 & 48.6 & 30.3 & 49.2 \\
{\footnotesize Llama-4-Maverick-17B (Text-Only)} & 11.0 & 45.8 & 7.3 & 46.8 & 7.6 & 46.8 & 8.5 & 46.5 \\
{\footnotesize Llama-4-Maverick-17B (Image-Only)} & 21.6 & 34.3 & 26.4 & 33.6 & 26.0 & 32.5 & 24.8 & 33.5 \\
\midrule
{\footnotesize Phi-3.5-Vision-instruct}                & 25.6 & 36.4 & 29.7 &34.9 & -- & -- & 27.7 & 36.2 \\
{\footnotesize Phi-3.5-Vision-instruct (Text-Only)}    & 16.7 & 37.6 & 17.7 & 38.3 & 17.7 & 38.7 & 17.4 & 38.2 \\
{\footnotesize Phi-3.5-Vision-instruct (Image-Only)}   & 11.8 & 25.4 & 9.6 & 21.1 & -- & -- & 10.7 & 23.2 \\
\midrule
{\footnotesize Phi-4-multimodal-instruct}             & \textbf{34.3} & 41.5 & \textbf{42.2} & 44.3 & \textbf{40.4} & 42.2 & \textbf{39.2} & 42.8 \\
{\footnotesize Phi-4-multimodal-instruct (Text-Only)} & 10.4 & 39.0 & 4.2 & 36.1 & 3.5 & 35.7 & 5.9 & 36.9 \\
{\footnotesize Phi-4-multimodal-instruct (Image-Only)}& 23.6 & 28.8 & 27.1 & 27.9 & 27.8 & 29.0 & 26.2 & 28.5 \\
\midrule
{\footnotesize Nemotron-Nano-12B-v2-vl}             & 11.2 & 38.6 & 6.6 & 38.1 & 6.2 & 37.3 & 8.4 & 38.1 \\
{\footnotesize Nemotron-Nano-12B-v2-vl (Text-Only)} & 13.0 & 42.7 & 6.5 & 41.9 & 6.4 & 41.5 & 8.5 & 42.0 \\
{\footnotesize Nemotron-Nano-12B-v2-vl (Image-Only)}& 8.2 & 31.4 & 8.2 & 31.1 & 8.5 & 31.7 & 8.2 & 31.3 \\
\bottomrule
\end{tabular}}

\caption{\textbf{Overall performance on \textsc{MulTivationBench}.} Results are aggregated across all four task types and reported across story lengths (Short, Medium, Long) and input modalities (Multimodal, Text-Only, Image-Only). Metrics are reported as EM (Exact Match) accuracy and F1 (Example-based F1) (\%) score. Bold indicates best; underline indicates second-best; ``--'' indicates that no score is reported because the model could not process the corresponding context-length subset due to image context window limitations.}
\label{tab:master_results}
\end{table*}

\subsection{Experimental Analysis}

\paragraph{Overall Performance}
Table~\ref{tab:master_results} reports the overall performance of several MLLMs aggregated across all four task types. Overall, \textsc{MultivationBench} is challenging for all evaluated models. The results also suggest a trade-off between F1 score and EM accuracy. Some models are better at retrieving plausible motivation sets, while others are more accurate at recovering the full gold label set. For example, Grok-4.1-Fast achieves the highest overall F1 score (55.0\%), whereas Phi-4-multimodal-instruct obtains the highest EM accuracy (39.2\%). Appendix~\ref{App:sub_common} reports a common-subset comparability check, whole-story confidence intervals, and pairwise tests. Taken together, these results indicate that current models often recover plausible motivation subsets without consistently identifying the complete gold label set.

\paragraph{Modality Ablation}
To assess the contribution of different input modalities, we compare model performance under three settings: \textit{Multimodal}, caption-enriched \textit{Text-Only}, and \textit{Image-Only}. The prompt variations are detailed in Table~\ref{tab:modality_prompts}. Table~\ref{tab:master_results} shows a consistent gap between Text-Only and Multimodal performance. Across several strong models, Text-Only inputs preserve a relatively large portion of the Multimodal F1 score, while EM accuracy drops much more sharply. For example, Gemini-3-Flash retains a relatively high Text-Only F1, but its EM accuracy declines from 35.7\% to 13.6\%. A similar F1--EM divergence appears across other strong models. This pattern suggests that textual narratives often provide enough information to recover partially correct motivation sets, but not enough to identify the exact gold label set. Visual information can therefore refine and revise an inferred motivation by supplying disambiguating evidence beyond text alone. Appendix~\ref{app:paired_modality} provides further paired whole-story analyses of these modality effects.

\begin{figure*}[t]
    \centering
    \includegraphics[width=\textwidth]{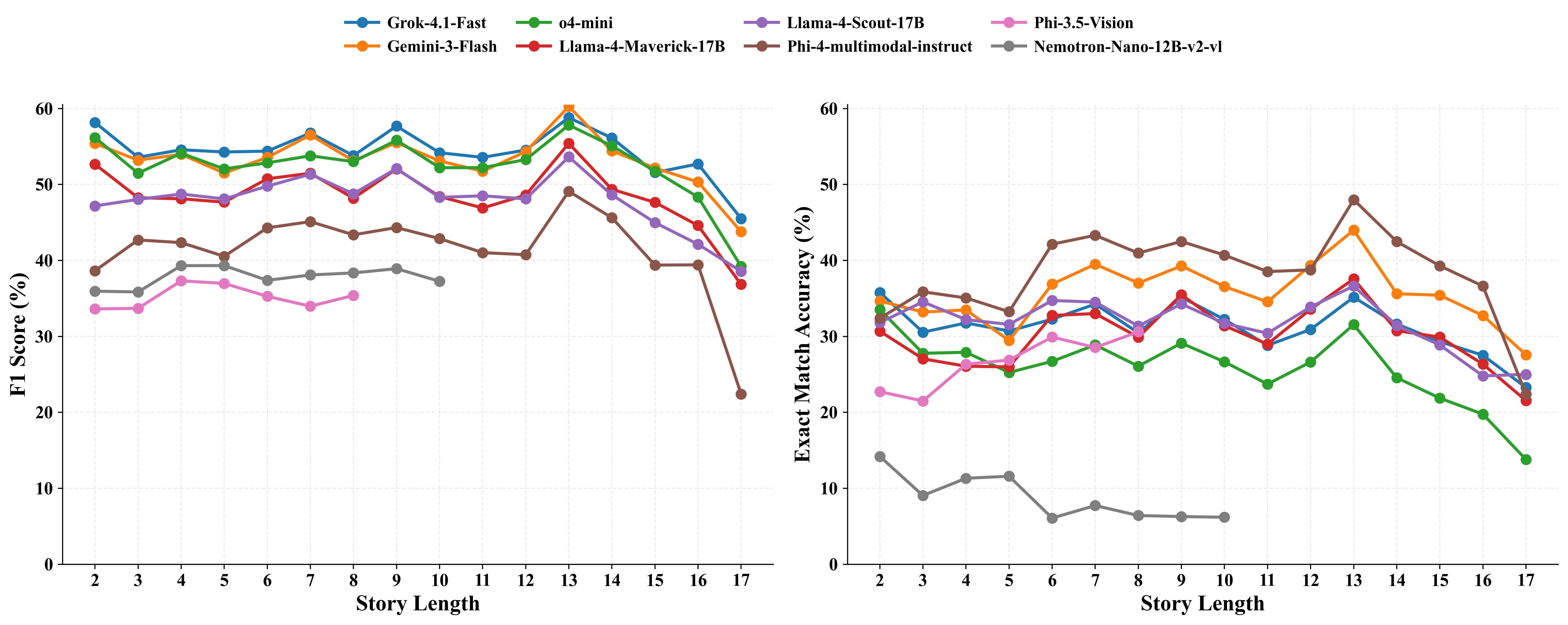}
    \caption{\textbf{Sequential Motivation Reasoning Performance.}
    Performance across increasing story lengths. Left: F1 score(\%). Right: Exact Match (EM) Accuracy(\%)}
    \label{fig:sequential_trend}
\end{figure*}

\paragraph{Impact of Narrative Length}
To study the effect of narrative length, we evaluate model performance across stories of different lengths. As shown in Table~\ref{tab:master_results}, most models improve from the Short to Medium interval, suggesting that current MLLMs can generally handle moderate multimodal contexts. However, this trend does not continue for longer narratives: for many models, performance in the Long setting declines relative to Medium. Figure~\ref{fig:sequential_trend} provides a finer-grained view. For story lengths between 2 and 13, performance remains relatively stable, while most models decline more consistently beyond length 13. The challenge of longer narratives is therefore not simply the presence of more context, but the need to trace and revise character motivations as new visual and textual evidence unfolds. Later evidence can change the most plausible explanation for behavior, requiring models to update their interpretation over the course of the story. Appendices~\ref{App:same_behavior_ablation} and \ref{app:adjacent_transitions} provide further same-behavior context-ablation and adjacent-transition analyses.

\paragraph{Story-level Consistency}
While the previous analyses examine instance-level performance under different conditions, we next test whether models maintain correct motivation reasoning across an entire story. Following \citet{chan2024negotiationtom}, we report a story-level \textbf{\textit{consistency score}}, which requires a model to answer every question within a story sequence correctly. As shown in Table~\ref{tab:story_consistency}, all models perform poorly on this metric. Even the best Full consistency score is only 0.80\%, indicating that current MLLMs rarely track character motivations correctly throughout an entire story. This weakness also appears in per-theory consistency: Phi-4-multimodal-instruct remains relatively balanced on Maslow (Definition: 15.4\%; Motivation: 15.2\%), whereas Llama-4-Maverick-17B shows a much larger gap (15.8\% vs.\ 2.7\%).

To address the concern that the all-correct criterion compresses model differences, Table~\ref{tab:story_consistency} further reports Story EM/F1. Each metric averages question-level scores within a story and then macro-averages across stories. Appendix~\ref{App:soft_story_metrics} provides the aggregation details. Story EM ranges from 8.65\% to 38.51\%, and Story F1 from 36.21\% to 54.80\%. These softer metrics preserve graded model differences, while Full serves as a high-stringency diagnostic of perfectly correct whole-story performance. These results show that models achieve meaningful partial success within stories, but this success rarely translates into fully correct performance across an entire narrative.

\begin{table*}[t]
\centering
\small
\setlength{\tabcolsep}{7pt}
\renewcommand{\arraystretch}{1.05}
\begin{tabular*}{\textwidth}{@{\extracolsep{\fill}}lrr|rrrrr@{}}
\toprule
\multirow{2}{*}{Model}
& \multicolumn{2}{c|}{Story-level scores}
& \multicolumn{2}{c}{Maslow}
& \multicolumn{2}{c}{Reiss}
& \multirow{2}{*}{Full} \\
\cmidrule(lr){2-3}\cmidrule(lr){4-5}\cmidrule(lr){6-7}
& EM & F1 & Def. & Mot. & Def. & Mot. & \\
\midrule
\multicolumn{8}{@{}l}{\textit{Closed-source}} \\
Grok-4.1-Fast  & 32.05 & \textbf{54.80} & 13.9 & 5.8 & \textbf{10.3} & 1.9 & 0.20 \\
Gemini-3-Flash & 35.48 & 53.74 & 11.2 & 9.7 & 9.8 & 3.4 & 0.50 \\
o4-mini        & 27.14 & 53.18 & 15.0 & 3.5 & 6.7 & 0.9 & 0.10 \\
\midrule
\multicolumn{8}{@{}l}{\textit{Open-source}} \\
Llama-4-Scout-17B           & 32.94 & 49.03 & 11.6 & 7.9 & 9.1 & 2.5 & 0.30 \\
Llama-4-Maverick-17B        & 29.86 & 49.02 & \textbf{15.8} & 2.7 & 9.5 & 1.5 & 0.10 \\
Phi-3.5-Vision-instruct     & 28.75 & 36.21 & 4.6 & 4.7 & 3.7 & 3.9 & 0.10 \\
Phi-4-multimodal-instruct   & \textbf{38.51} & 42.61 & 15.4 & \textbf{15.2} & 6.4 & \textbf{4.1} & \textbf{0.80} \\
Nemotron-Nano-12B-v2-vl     & 8.65 & 38.22 & 2.9 & 0.2 & 0.6 & 0.1 & 0.00 \\
\bottomrule
\end{tabular*}
\caption{\textbf{Story-level performance (\%).} Definition, Practical, and Full are all-correct consistency rates; Full spans all four tasks. Story EM/F1 first average question-level scores within each story and then macro-average across stories. Bold marks the column best.}
\label{tab:story_consistency}
\end{table*}

\subsection{Task-based Analysis}

\paragraph{Reasoning Task Comparison}
To analyze how MLLMs perform under different formulations of motivation reasoning, we compare results on \textit{Definition} tasks (theoretical matching) and \textit{Practical Motivation} tasks (multimodal inference). Table~\ref{tab:task_breakdown} reveals a model-dependent divergence between the two settings rather than a uniform difficulty gap. Some models perform substantially better on \textit{Definition} than on \textit{Practical Motivation}. For example, o4-mini drops from 44.1\% EM accuracy on Maslow \textit{Definition} tasks to 24.7\% on Maslow \textit{Practical Motivation} task, while Llama-4-Maverick-17B declines from 47.3\% to 21.7\%. This pattern suggests that these models retain substantial familiarity with psychological category definitions, but are less effective at applying those concepts to situated multimodal narratives. In contrast, other models show the reverse trend. Phi-4-multimodal-instruct improves from 47.5\% EM accuracy on Maslow \textit{Definition} tasks to 52.9\% on Maslow \textit{Practical Motivation} tasks, while Gemini-3-Flash increases from 33.1\% to 43.2\%. For these models, narrative context appears to provide useful grounding cues rather than introducing additional difficulty. This contrast suggests that \textit{Definition} and \textit{Practical Motivation} tasks probe partially distinct capabilities: semantic familiarity with psychological categories versus the ability to ground them in sequential multimodal narratives.

\begin{table}[t]
\centering
\scriptsize
\setlength{\tabcolsep}{2.6pt}
\renewcommand{\arraystretch}{0.98}
\resizebox{\columnwidth}{!}{\begin{tabular}{@{}l|cc|cc|cc|cc@{}}
\toprule
\multirow{3}{*}{Model}
& \multicolumn{4}{c|}{Maslow (8)}
& \multicolumn{4}{c@{}}{Reiss (16)} \\
\cmidrule(lr){2-5} \cmidrule(l){6-9}
& \multicolumn{2}{c|}{Def.}
& \multicolumn{2}{c|}{Mot.}
& \multicolumn{2}{c|}{Def.}
& \multicolumn{2}{c@{}}{Mot.} \\
\cmidrule(lr){2-9}
& EM & F1 & EM & F1 & EM & F1 & EM & F1 \\
\midrule
\multicolumn{9}{@{}l}{\textit{Closed-source}} \\
Grok-4.1-Fast              & 41.9 & \textbf{63.3} & 31.2 & \textbf{63.2} & 37.2 & \textbf{48.3} & 17.2 & 45.1 \\
Gemini-3-Flash             & 33.1 & 58.9          & 43.2 & 62.8          & 36.0 & 47.2          & \textbf{30.7} & \textbf{46.0} \\
o4-mini                    & 44.1 & 62.6          & 24.7 & 61.1          & 26.9 & 47.8          & 10.8 & 41.0 \\
\midrule
\multicolumn{9}{@{}l}{\textit{Open-source}} \\
Llama-4-Scout-17B          & 37.6 & 58.9          & 37.7 & 58.3          & 36.4 & 42.0          & 19.0 & 37.0 \\
Llama-4-Maverick-17B       & 47.3 & 59.1          & 21.7 & 54.8          & \textbf{38.3} & 46.3 & 14.0 & 36.7 \\
Phi-3.5-Vision-instruct    & 26.3 & 31.2          & 35.1 & 54.0          & 20.4 & 24.0          & 28.9 & 33.4 \\
Phi-4-multimodal-instruct  & \textbf{47.5} & 51.0 & \textbf{52.9} & 57.7 & 28.6 & 31.4 & 28.0 & 31.2 \\
Nemotron-Nano-12B-v2-vl    & 22.2 & 45.1          & 4.7  & 42.7          & 3.9  & 33.2          & 2.9  & 31.9 \\
\midrule
Human                      & 70.6 & 81.5          & 78.6 & 87.6          & 60.7 & 74.2          & 63.9 & 72.9 \\
\bottomrule
\end{tabular}}
\caption{\textbf{Task-level performance on \textsc{MULTIVATIONBENCH}.}
EM accuracy and F1 score (\%) for Maslow (8) and Reiss (16) across Definition and Practical Motivation tasks.
Bold indicates the best \emph{model} result in each column.}
\label{tab:task_breakdown}
\end{table}

\paragraph{Taxonomy Granularity}
A clear overall pattern in Table~\ref{tab:task_breakdown} is the performance drop from the coarser Maslow taxonomy to the more fine-grained Reiss taxonomy, especially in the \textit{Practical Motivation} setting. For example, Grok-4.1-Fast achieves 63.2\% F1 score on Maslow \textit{Practical Motivation}, but falls to 45.1\% on the corresponding Reiss task. Similar declines can be observed across most models, indicating that current MLLMs can often identify a broad motivational region but struggle when the task requires finer discrimination among closely related psychological alternatives. This pattern suggests that fine-grained motivation reasoning remains a major challenge even when models achieve comparatively strong results on coarser taxonomies.

\paragraph{Human Comparison}
Human performance remains substantially above all evaluated models across task settings. As shown in Table~\ref{tab:task_breakdown}, the gap is especially large on the Practical Motivation tasks. For example, in the Reiss \textit{Practical Motivation} setting, human Exact Match reaches 63.9\%, compared with 30.7\% for the best-performing model (Gemini-3-Flash). A similarly large gap is observed in F1 score, with human performance reaching 72.9\% compared with 46.0\% for the strongest model. These results indicate that \textsc{MulTivationBench} is far from saturated and that current MLLMs remain substantially limited in grounded and fine-grained sequential motivation reasoning.

\begin{figure*}[t]
    \centering
    \includegraphics[width=1\textwidth]{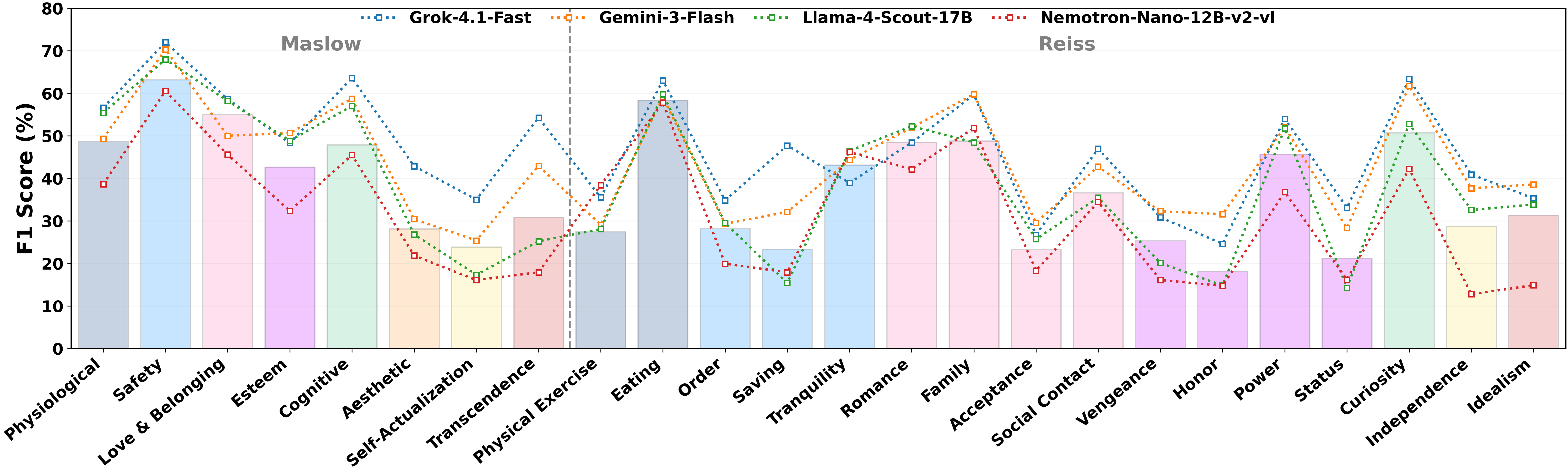}
    \caption{Label-wise F1 across the eight Maslow need categories (left) and the 16 Reiss desire categories (right) options in the \textsc{MulTivationBench}. Bars show the mean F1 score aggregated across all evaluated models for each option; overlaid lines show per-option trajectories for the selected representative models.}
    \label{fig:label_performance}
\end{figure*}

\begin{figure}[t]
    \centering
    \includegraphics[width=\linewidth]{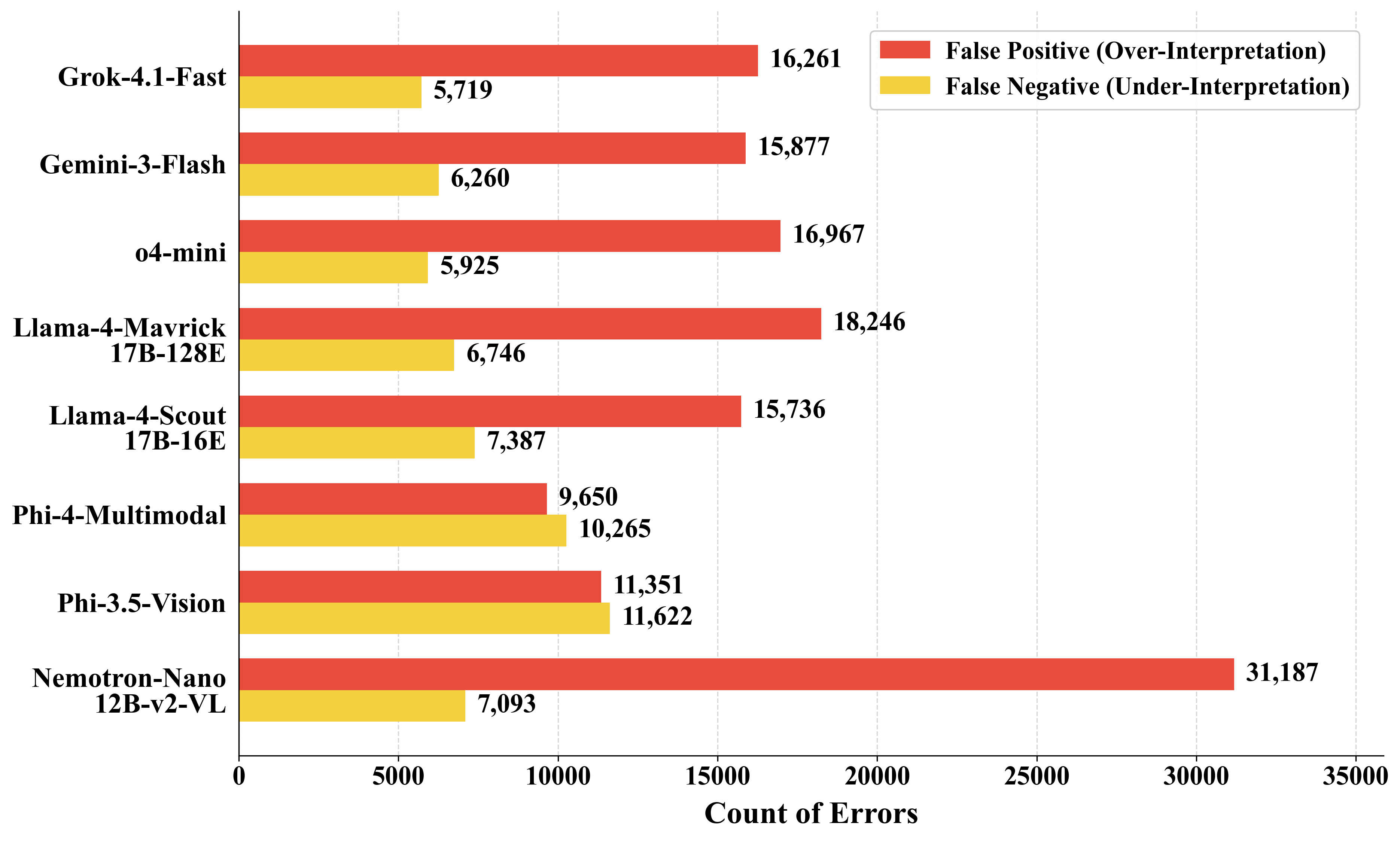}
    {}
    \caption{Grouped bar chart comparing the frequency of False Positive (Red) versus False Negative (Yellow) errors among all models.}
    \label{fig:error_counts}
\end{figure}

\subsection{How MLLMs Make Errors}
\label{sec:error_analysis}
To further understand why performance degrades as story context grows longer, we analyze the error patterns of MLLMs. Following \citet{lee2025mindmotions}, we define \textit{Over-Interpretation} (False Positive) as instances where a model hallucinates a motivation that is not supported by the ground truth, and \textit{Under-Interpretation} (False Negative) as instances where it fails to recognize a valid motivation.

\paragraph{Over- vs. Under-interpretation} As illustrated in Figure~\ref{fig:error_counts}, we observe a consistent bias toward Over-Interpretation across the majority of high-capacity models. Notably, Nemotron-Nano-12B-v2-vl exhibits the largest disparity, committing over 31,000 False Positive errors compared to approximately 7,000 False Negatives. This suggests that as story information accumulates, the model struggles to discard earlier but less relevant cues when inferring a character's current motivation, leading to substantial over-prediction. Similarly, closed-source models such as Grok-4.1-Fast and Gemini-3-Flash also show a strong False Positive bias, with nearly three times more False Positives than False Negatives. These results suggest that the main failure mode of current MLLMs is not simply overlooking valid motivations, but failing to revise earlier interpretations as new evidence unfolds.

\paragraph{Label-wise Performance on Desire and Need Options}
To further examine performance variation across motivation labels, Figure~\ref{fig:label_performance} reports the average F1 score for each label across models. A clear pattern is that MLLMs perform better on labels that can be inferred from a single salient cue or local scene, such as \textit{Eating}, \textit{Safety}, and \textit{Romance} reaching from 50\% to 60\%. These labels are often supported by direct evidence in one moment, such as eating food, reacting to danger, or a couple kissing. In contrast, performance drops on labels such as \textit{Self-Actualization}, \textit{Transcendence}, \textit{Status}, and \textit{Independence} reaching from 20\% to 30\%, whose interpretation typically requires accumulated context about the character's longer-term goals, abstracted values, or social position. This trend suggests that MLLMs can handle explicit scene-level motivation cues, but struggle to identify correct abstracted motivations that require information accumulate across the narrative.

\section{Conclusion}
\label{sec:conclusion}

We introduce \textsc{MulTivationBench}, a benchmark for multimodal sequential motivation reasoning grounded in Maslow's hierarchy and Reiss's basic desires. Results show that current MLLMs identify locally plausible motivations but struggle with consistent reasoning across full narratives.

\section{Limitations}
\label{sec:limitations}
\paragraph{Passive benchmark to evaluate the ability of LLMs}
While \textsc{MulTivationBench} advances the evaluation of multimodal sequential motivation reasoning, it remains an offline benchmark that evaluates models as passive observers of completed narratives~\citep{chan2024negotiationtom,ma-etal-2023-towards-holistic}. In our current setting, models infer motivations from accumulated visual and textual context, but they are not required to act, intervene, or predict how a character's motivation may change under alternative future observations or environmental changes. The active reasoning benchmark should treat the language model as an active agent that perceives the physical and social context, reasons about others' mental states, communicates with other agents, and interacts with the environment to complete predefined tasks~\citep{yim2024guandantom}. A natural direction for future work is therefore to extend this benchmark toward world-model-based motivation reasoning. Rather than only asking why a behavior occurred, future evaluations could test whether a model can anticipate how motivations evolve, update earlier inferences after new evidence, and simulate how different interventions may alter a character's future behavior. We believe this is an important next step toward more active and embodied forms of social intelligence. Because Grok and Gemini participated in benchmark construction and are also evaluated, stylistic or distributional alignment may still \mbox{affect} their results. Appendix~\ref{app:generator-bias} reports a robustness analysis, but it cannot rule out all such effects. Similarly, the option controls in Appendix~\ref{appendix:options} do not exclude combinations of lexical cues.

\section{Ethics Statement}

In this work, we followed applicable data usage policies and took steps to minimize privacy, licensing, and content-related risks. \textsc{MulTivationBench} is built upon three source datasets: MovieBench (CC BY 4.0), StoryReasoning (CC BY-ND 4.0), and SSID (CC BY-NC-ND 4.0). For StoryReasoning and SSID, which carry ``No-Derivatives'' restrictions, we obtained explicit written permission from the respective dataset authors to release the derived character-motivation and behavior-focused Q\&A annotations generated in this work. In accordance with these permissions, our release contains only newly generated derived annotations, source identifiers, reconstruction guidance, evaluation code, and scripts or documentation for obtaining the corresponding images and story context from upstream datasets. We do not redistribute original images, videos, story texts, or any substantial portion of the source datasets. The SSID-derived subset is restricted to non-commercial academic research and evaluation, consistent with the granted terms. All released materials clearly attribute the corresponding original datasets and publications. During curation, we filtered out candidate stories and samples containing potentially offensive or inappropriate narrative content. Because the benchmark is derived from fictional stories and publicly available media, and because we do not redistribute the original multimodal content, we do not anticipate material privacy risks to real individuals from the released materials. Appendix~\ref{app:release_reproducibility} gives the complete release boundary and reconstruction workflow.

The benchmark should not be used to infer real people's motives in surveillance, employment, education, hiring, policing, or other high-stakes settings. Predictions are uncertain, context-dependent hypotheses and must not be presented as factual psychological assessments. Research use should favor calibrated or uncertainty-aware outputs and retain meaningful human oversight.

\section{Acknowledgments}

The authors of this paper were supported by the National Key Research and Development Program of China (2025YFE0200500), the ITSP Platform Research Project (ITS/189/23FP) from ITC of Hong Kong, SAR, China, and the AoE (AoE/E-601/24-N), the CRF (No. C6004-25G),  the RIF (R6021-20) and the GRF (16205322) from RGC of Hong Kong, SAR, China. We also thank the support from Amazon. 

\bibliography{custom}

\appendix

\section{Appendix for the Motivation Frameworks}
\label{appendix:A}

\subsection{Choice of Psychological Frameworks}
\label{app:framework_choice}

We use Maslow's expanded hierarchy and Reiss's 16 basic desires as complementary operational taxonomies rather than as an exhaustive theory of human motivation. Maslow provides a coarse-grained need-level structure covering broad categories such as safety, belonging, esteem, cognitive needs, self-actualization, and transcendence. This makes it suitable for evaluating whether models can identify the broad motivational region expressed by a behavior. Reiss's theory provides a finer-grained desire-level taxonomy, distinguishing motivations such as family, acceptance, social contact, independence, status, curiosity, and idealism. This enables a more detailed evaluation of whether models can discriminate among closely related motivational contents.

This design supports the central coarse-to-fine comparison in our benchmark. The Maslow tasks evaluate broad need recognition, while the Reiss tasks evaluate fine-grained desire discrimination. We do not assume that Maslow's hierarchy is the only valid or definitive theory of motivation, nor do we rely on a strong claim that needs must always be satisfied in a strict hierarchy. Instead, we use the expanded hierarchy as an interpretable broad-level label space.

Other theories of motivation are also relevant. For example, Self-Determination Theory \citep{ryan2000self} emphasizes autonomy, competence, and relatedness as central dimensions of intrinsic motivation. These dimensions are useful for studying intrinsic motivation, but they provide a smaller and more abstract label space than required for our current multi-label benchmark over diverse everyday narrative behaviors. We therefore view Self-Determination Theory as a valuable direction for future extensions of \textsc{MultivationBench}, especially for studying intrinsic motivation in greater detail.

\label{sec:appendix_framework_details}

\subsection{Details of the Hierarchy of Needs}
Maslow's Expanded Hierarchy of Needs~\citep{maslow1970motivation} is a motivational theory that explains how human needs may be prioritized and fulfilled, from basic survival to higher-order psychological growth and meaning.
While Maslow originally discussed five levels, later formulations commonly expand the hierarchy; in this work we adopt an 8-level version (Physiological, Safety, Love \& Belonging, Esteem, Cognitive, Aesthetic, Self-Actualization, Transcendence) to cover both lower-level and higher-level motives in a unified taxonomy.

\noindent\textbullet\ \textbf{Physiological Needs:}
Physiological needs refer to essential survival requirements such as food, water, warmth, and sleep.
These needs form the foundation of the hierarchy and must be sufficiently met before individuals can consistently pursue higher-level goals.

\noindent\textbullet\ \textbf{Safety Needs:}
Safety needs emphasize stability and protection, including physical security, health, order, and freedom from fear.
They capture motivations related to predictability and safeguarding one's future.

\noindent\textbullet\ \textbf{Love and Belonging Needs:}
Love and belonging needs reflect the drive for social connection through friendship, intimacy, trust, and affiliation with groups (e.g., family, peers, community).
They involve motivations to be accepted, supported, and meaningfully connected to others.

\noindent\textbullet\ \textbf{Esteem Needs:}
Esteem needs involve both internal self-worth (e.g., competence, independence, achievement) and external recognition (e.g., respect, status, prestige).
Meeting these needs supports confidence and social standing.

\noindent\textbullet\ \textbf{Cognitive Needs:}
Cognitive needs reflect motivations for knowledge and understanding, including curiosity, exploration, and the pursuit of meaning and predictability.

\noindent\textbullet\ \textbf{Aesthetic Needs:}
Aesthetic needs involve appreciation and pursuit of beauty, balance, and form, capturing motivations oriented toward harmony and aesthetic experience.

\noindent\textbullet\ \textbf{Self-Actualization Needs:}
Self-actualization concerns realizing personal potential through growth, mastery, and self-fulfillment.
It is often expressed via creativity, goal pursuit, and striving toward an ideal self.

\noindent\textbullet\ \textbf{Transcendence Needs:}
Transcendence needs describe motivations beyond the personal self, such as altruism, spiritual connection, and contributing to a larger purpose or the well-being of others.

\subsection{Details of Reiss's Basic Desires and the RMP}
The Reiss theory of 16 basic desires~\citep{reiss2004} provides a complementary view of intrinsic motivation, positing that behavior can be explained by preferences over 16 fundamental motivational drives.
The Reiss Motivation Profile (RMP) operationalizes this idea as an inventory: individuals differ in the intensity of each desire, and these differences shape behavior, decision-making, and lifestyle.

To support interpretability when discussing fine-grained motivation types, we summarize the 16 desires using Maslow-style groupings.
\textbf{Physiological}-oriented desires such as \textit{Eating} and \textit{Physical Exercise} capture bodily drives and activity.
\textbf{Safety}-oriented desires such as \textit{Order}, \textit{Saving}, and \textit{Tranquility} reflect preferences for structure, safeguarding one's future, and peace of mind.
\textbf{Love \& Belonging}-oriented desires include \textit{Acceptance}, \textit{Family}, \textit{Romance}, and \textit{Social Contact}, spanning approval, close bonds, companionship, and affiliation.
\textbf{Esteem}-oriented desires such as \textit{Status}, \textit{Power}, \textit{Vengeance}, and \textit{Honor} emphasize recognition, prestige, influence, and status-defense.
\textbf{Cognitive}-oriented desires such as \textit{Curiosity} motivate knowledge seeking, understanding, and exploration.
\textbf{Self-Actualization}-oriented desires such as \textit{Independence} emphasize autonomy, mastery, and personal growth.
Finally, \textbf{Transcendence}-oriented desires such as \textit{Idealism} capture value-driven motives beyond the personal self, including prosocial principles and contributing to a larger purpose.

\section{Appendix for \textsc{MultivationBench}}

\subsection{Verification of Potential Contamination}
\label{sec:contamination}

A potential concern is that some source materials underlying \textsc{MultivationBench} may have been exposed to large language models during training, since widely used NLP and multimodal benchmarks are often incorporated into both pre-training and post-training pipelines \citep{GolchinS24}. This issue is particularly relevant because \textsc{MultivationBench} integrates visual narratives from diverse sources, including MovieBench, SSID, and StoryReasoning, whose root materials may partially overlap with publicly available web data.

To assess this risk, we follow the \textit{Slot Guessing for Perturbed Caption} methodology proposed by \citet{song-etal-2025-text}. For each constituent dataset, we randomly sample 100 instances and evaluate the model under two conditions: (1) an \textbf{Original} condition using the original story text, and (2) a \textbf{Perturbed} condition using meaning-preserving paraphrases. We then report the \textit{Performance Gap} ($\Delta$) between the two conditions and the \textit{Memorization Ratio} ($\Phi$). Under this diagnostic, a substantially negative $\Delta$ would indicate stronger reliance on exact surface forms, and thus provide evidence more consistent with memorization than robust reasoning.

As shown in Table~\ref{tab:contamination}, the observed performance gaps are small across all three source datasets. MovieBench exhibits the largest negative gap ($\Delta=-0.07$), suggesting a limited contamination signal, which is plausible given the prevalence of movie-related text in public training corpora. SSID shows only a minor negative gap ($\Delta=-0.03$). By contrast, StoryReasoning yields a positive gap ($\Delta=+0.06$), indicating that model performance is not tied to the original wording and showing no evidence of strong phrase-level memorization under this test.

While no systematic approach can fully rule out contamination without access to private training sets, we include this analysis to explicitly examine the risk and to verify that benchmark performance is not solely driven by exact-match recall. Overall, the results suggest that \textsc{MultivationBench} is not predominantly exploiting memorized textual patterns, and instead serves primarily as a test of reasoning over visual narratives.

\begin{table}[h]
\centering
\resizebox{\columnwidth}{!}{\begin{tabular}{lccc}
\toprule
\textbf{Dataset Source} & \textbf{$\Delta$ (Gap)} & \textbf{$\Phi$ (Mem. \%)} & \textbf{Risk Level} \\
\midrule
MovieBench & -0.07 & 22.0\% & Low \\
SSID & -0.03 & 17.5\% & Low \\
StoryReasoning & +0.06 & 25.0\% & Very Low \\
\bottomrule
\end{tabular}}
\caption{\textbf{Contamination probe results.} We compare model performance under original and paraphrased story text. More negative $\Delta$ values indicate greater sensitivity to exact wording and therefore stronger evidence consistent with memorization. Overall, all three \textsc{MultivationBench} sources exhibit only small performance gaps, suggesting limited contamination signals under this diagnostic.}
\label{tab:contamination}
\end{table}

\subsection{Analysis of Potential Generator-Family Advantage}
\label{app:generator-bias}

\begin{table}[t]
\centering
\small
\setlength{\tabcolsep}{3.5pt}
\begin{tabular}{llcc}
\toprule
Statistic & Metric & Estimate & 95\% CI / $p$ \\
\midrule
Overall $\Delta$ & EM accuracy & +5.88 & [5.29, 6.46] \\
Overall $\Delta$ & F1 score & +9.42 & [8.94, 9.89] \\
Gap $\Delta_g$ & EM accuracy & +0.77 & [-0.42, 1.95] \\
Gap $\Delta_g$ & F1 score & -1.99 & [-2.94, -1.02] \\
Perm. test on $\Delta_g$ & EM accuracy & +0.77 & $p=0.93$ \\
Perm. test on $\Delta_g$ & F1 score & -1.99 & $p=0.79$ \\
\bottomrule
\end{tabular}
\caption{Generator-family analysis in the main \texttt{Multimodal} setting.
Overall $\Delta$ is the difference between the mean performance of the generator-family and non-generator-family groups. 
$\Delta_g = (\text{Practical}-\text{Definition})_{\text{gen}} - (\text{Practical}-\text{Definition})_{\text{non-gen}}$
measures whether the generator-family models obtain an additional advantage specifically on the \textit{Practical} tasks.
While generator-family models perform better overall, the gap analysis does not indicate a corresponding Practical-specific benefit.}
\label{tab:generator_bias_bootstrap}
\end{table}

A natural concern is that overlap between the benchmark construction pipeline and some evaluated model families may favor those models at test time. This concern can be split into two separate questions. First, are generator-family models simply stronger overall? Second, do they enjoy an additional advantage specifically on the \textit{Practical} tasks, where overlap with the construction pipeline is most plausible?

To address this, we examine both the overall group difference and the difference in the \textit{Practical}-vs.-\textit{Definition} performance gaps:
\[
\begin{aligned}
\Delta_g
&=
(\text{Practical}-\text{Definition})_{\text{gen}} \\
&\quad -
(\text{Practical}-\text{Definition})_{\text{non-gen}} .
\end{aligned}
\]
Under a strong family-specific construction bias, we would expect not only a positive overall difference, but also a clearly positive $\Delta_g$ in the main \texttt{Multimodal} multimodal setting.

We estimate uncertainty with nonparametric bootstrap resampling, following standard statistical practice in NLP evaluation~\citep{efron1979,dror-etal-2018-hitchhikers,berg-kirkpatrick-etal-2012-empirical,yeh-2000-accurate}. We also run an exact permutation test over model-family assignments. The results are shown in Table~\ref{tab:generator_bias_bootstrap}.

The analysis reveals a clear overall advantage for generator-family models in the main multimodal setting. The mean difference between the generator-family and non-generator-family groups is +5.88 points in EM accuracy (95\% CI: [5.29, 6.46]) and +9.42 points in F1 score (95\% CI: [8.94, 9.89]). However, this overall advantage does not extend to a reliable extra gain on the \textit{Practical} tasks. For EM accuracy, the estimated gap difference is small (+0.77), and its 95\% confidence interval crosses zero ([-0.42, 1.95]). For F1 score, the estimated gap difference is negative (-1.99), with a 95\% confidence interval below zero ([-2.94, -1.02]).

The permutation test leads to the same conclusion. The observed \textit{Practical}-vs.-\textit{Definition} gap difference is not unusual under random family assignment (EM accuracy: $p=0.93$; F1 score: $p=0.79$). In other words, the actual generator-family grouping does not show an unusually large Practical-specific effect.

Overall, these results suggest that generator-family models are stronger baselines in general, but they do not provide evidence for an additional family-specific advantage on the model-generated \textit{Practical} tasks in the main multimodal setting. Construction/evaluation overlap remains a reasonable caveat, but our analysis does not indicate that it is a major driver of the benchmark results through a robust Practical-specific effect. We do not claim that this fully rules out all forms of stylistic or distributional alignment; rather, it shows that such an effect is not supported by the present gap-based analysis.
\section{Appendix for Related Works}
\label{Appendix_for_Related_Works}

\subsection{Related Works for Large language
Models}
\label{Appendix_for_Related_Work_LLMs}

Recent studies have demonstrated the remarkable capabilities of instruction-following large language models (LLMs)~\cite{openai2023gpt4technical, openai_o4mini_2025, google2025gemini3flash, xai2025grok41, jiang2023lion}, showing strong zero-shot and few-shot performance across a wide range of natural language processing tasks~\cite{
bubeck2023sparks,
chan-etal-2024-exploring,
jiayang-etal-2023-storyanalogy,
wang-etal-2024-candle,
jiayang-etal-2024-econ,
shi-etal-2026-inferencedynamics,
jiayang-etal-2024-eventground, chan-etal-2023-self}. Despite these impressive advances, existing studies have identified several reasoning challenges that remain difficult for current LLMs, including complex mathematical reasoning~\cite{frieder2023mathematical}, theory of mind reasoning~\cite{lin2024constrainedreasoning}, uncertainty and confidence calibration~\cite{zong2025critical}, retrieval-augmented generation~\cite{jiayang-etal-2025-integround,zhang2025agenticdeepresearch,li-etal-2025-survey}, intention reasoning~\cite{yang2026sessionintentbench}, analogical reasoning~\cite{jiayang-etal-2023-storyanalogy}, discourse relation classification~\cite{chan-etal-2023-discoprompt}, text-to-table generation~\cite{deng-etal-2024-text,deng2025structuring}, complex game scenarios~\cite{yim2024guandantom,mo-etal-2026-dixitworld,liang2025llmhanabi}, argument impact classification~\cite{chan2024audiencepersona}, and the associated ethical and privacy challenges~\cite{li2023llmprivacy,susnjak2024examintegrity,li-etal-2024-privlm,lukas2023piileakage,li2025simulateeliminate}.
State-of-the-art LLMs, including o4-mini~\citep{openai_o4mini_2025}, Gemini-3-Flash~\citep{google2025gemini3flash}, Grok-4.1-Fast~\citep{xai2025grok41}, Llama-4 family (Scout-17B and Maverick-17B)~\citep{meta2025llama4}, and DeepSeek~\cite{deepseekai2025deepseekr1incentivizingreasoningcapability}, have significantly improved reasoning performance through large-scale pre-training and reinforcement learning. Nevertheless, existing evaluations have primarily focused on static reasoning tasks, isolated text understanding, or single-step multimodal perception. Relatively little attention has been devoted to evaluating whether these models can infer and continuously update human motivations as additional multimodal evidence unfolds throughout a narrative. Since human motivation is an inherently latent and dynamic mental state, effective reasoning requires not only recognizing observable behaviors but also integrating accumulated visual and textual context to revise previous interpretations. Our work complements existing LLM evaluation benchmarks by introducing a new dimension of social reasoning: multimodal sequential motivation reasoning. Unlike previous benchmarks that emphasize static social commonsense, intention prediction, or theory-of-mind reasoning, MULTIVATIONBENCH evaluates whether models can consistently infer psychologically grounded motivations under evolving multimodal contexts.

\subsection{Related Works for Multimodal Social and Theory-of-Mind Reasoning}
\label{sec:appendix_related_multimodal_social_tom}

Theory-of-mind (ToM) reasoning is a central testbed for evaluating whether models can infer others' latent mental states, including beliefs, desires, intentions, emotions, and goals. Prior benchmarks have substantially advanced this direction through information-asymmetric conversations, cross-cultural ToM, negotiation-based interactions, and systematic social-cognitive evaluations~\citep{kim2023fantom, XToM2025, chan2024negotiationtom, tombench2025}. However, these works mainly focus on text-only settings or structured interaction scenarios.

Recent multimodal social-reasoning benchmarks extend evaluation beyond text-only stories by integrating visual evidence with social context. MMToM-QA evaluates goals, beliefs, emotions, and social relations in video~\citep{jin2024mmtomqa}; Social Genome evaluates grounded reasoning traces over visual, verbal, vocal, and external-knowledge evidence~\citep{mathur-etal-2025-social}; and SIV-Bench covers social-scene understanding, social-state reasoning, and social-dynamics prediction~\citep{kong2025sivbench}. Complementary benchmarks study visual social commonsense~\citep{lin2025valphasocial}, realistic short-film ToM~\citep{villacueva2025moments}, mime-based nonverbal reasoning~\citep{li2025mimeqa}, body language~\citep{lee2025mindmotions}, multi-person gestures~\citep{Cao_2025_CVPR}, grounded physical-social norms~\citep{rezaei-etal-2025-egonormia}, and densely aligned multimodal interactions~\citep{Lee_2024_CVPR}. These works establish the importance of multimodal social cues but do not evaluate Maslow- and Reiss-grounded motivation labels as narrative evidence accumulates.

\subsection{Related Works for Temporal Visual Understanding and Long-Video Reasoning}
\label{sec:appendix_related_temporal_video}

Another line of work evaluates temporally extended visual inputs. Zhu et al.~\citep{zhu2023sequential} benchmark future-event reasoning from sequential images across abstract patterns, human activities, and physical interactions. MVBench, EgoSchema, Video-Bench, Video-MME, LongVideoBench, and MLVU test temporal perception, long-form video-language understanding, and multi-task video reasoning~\citep{li2024mvbench,mangalam2023egoschema,ning2023videobench,fu2024videomme,wu2024longvideobench,zhou2025mlvu}. Video-ChatGPT, TimeChat, and Temporal Grounding Bridge further study video conversation, time-sensitive understanding, and temporal extrapolation~\citep{maaz2024videochatgpt,ren2024timechat,wang2024tgb}. Their central questions concern what happens, when it happens, or what happens next, rather than taxonomy-grounded explanations of why a character acts.

\subsection{Related Works for Visual Storytelling and Narrative Consistency}
\label{sec:appendix_related_visual_storytelling}

Visual storytelling research studies how models generate or maintain coherent narratives across images and videos. StoryDALL-E adapts pretrained text-to-image transformers for story continuation~\citep{maharana2022storydalle}, and ViSTA uses multimodal adapters to condition text-to-image diffusion on visual storytelling history~\citep{dong2025vista}. StoryDiffusion targets long-range consistency for image and video generation through consistent self-attention~\citep{zhou2024storydiffusion}, while Storynizor and CharaConsist focus on character-consistent story generation and fine-grained character consistency~\citep{ma2025storynizor,wang2025characonsist}. MovieBench provides a hierarchical movie-level dataset for long-video generation, further reflecting the need for narrative structure and long-range visual coherence~\citep{wu2024moviebench}. These works are relevant because they address long-range stories, character continuity, and visual narrative coherence. Their central objective, however, is generation or consistency maintenance, whereas our benchmark evaluates recognition and reasoning over existing visual narratives.

\subsection{Position of \textsc{MultivationBench}}
\label{sec:appendix_multivationbench_position}

Prior work therefore covers three adjacent capabilities: multimodal social reasoning, temporal video understanding, and coherent visual storytelling. \textsc{MultivationBench} connects these directions but evaluates a different target capability. It requires models to ground behavior in multimodal evidence, accumulate sequential narrative context, and assign psychologically grounded motivation labels from Maslow's hierarchy and Reiss's basic desires. This setting asks not only what a character did or when an event occurred, but why the behavior is best explained by a particular motivation after the story context has accumulated. In this sense, \textsc{MultivationBench} complements existing benchmarks by isolating sequential motivation reasoning as a distinct form of multimodal social intelligence.

\section{Data Construction Pipeline}
\label{Appendix:DataConstruction}
\subsection{Detailed Data Pre-processing and Prompts} \label{Appendix:pre} To curate meaningful evaluation samples, we focused on visually grounded and psychologically salient character behaviors, rather than background-only presence or trivial scene details. We automated this filtering process using a multi-model extraction pipeline designed to minimize single-model bias~\citep{gu2026llmasajudge}. Two state-of-the-art MLLMs, Grok-4.1 and Gemini-3, independently scanned the images and story context to identify the main character and extract the major \textbf{Character} and \textbf{Behavior Chain}---a sequence of \textbf{Behaviors} mapped to specific image indices with hypothesized \textbf{Motivations}.

Subsequently, a reasoning-specialized model (Grok-4.1-fast-reasoning) acted as an adjudicator, reviewing the candidate outputs to select the most coherent chain that maintained strict temporal logic and character consistency. This consensus-based approach reduces dependence on a single generator~\citep{HuangYMZFWCPFQL25}. Finally, the automated outputs underwent manual review using three criteria: \begin{itemize} \item \textbf{Visual Grounding:} Ensuring the described cue is clearly visible in the image frame. \item \textbf{Psychological Salience:} Verifying that the inferred motivation is relevant to the scene rather than trivial movement. \item \textbf{Human Character Focus:} Confirming that the chain consistently tracks the main character without drifting to background characters. \end{itemize}

The stored curation trail begins with 1{,}235 automatically screened valid stories, from which 1{,}000 story slots were selected. The original question-generation branch contains 4{,}275 behavior candidates; 32 story slots then underwent replacement or character-chain correction, producing 107 replacement candidates. Counting both batches gives 4{,}382 candidates, of which 4{,}023 remain after filtering and final allow-listing. A separate repair pass rewrote 101 question records. These counts distinguish calibration, filtering, targeted correction, and question regeneration from the final benchmark size.

\vspace{0.5em} \paragraph{Prompt Templates} We present the specific extractor and reviewer prompts used in this stage in Tables \ref{tab:extractor_prompt} and \ref{tab:reviewer_prompt}. The extractor template is designed to identify the main character, behavior, and motivation given a sequence of images and the full story context, outputting the structured Behavior Chain. The reviewer prompt evaluates these generated outputs to select the optimal chain based on the criteria described above.

\subsection{Details of Options Generation Protocol} \label{appendix:options} Existing benchmarks mostly rely on manual construction, which is labor-intensive and limits scalability. To reduce financial and labor costs while ensuring theoretical rigor, an Automated Multi-Stage framework was employed to formulate questions and generate options, as illustrated in Figure \ref{fig:pipeline}. The detailed algorithm is provided in Algorithm \ref{alg:option_gen}.

We focused our generation efforts on the \textit{Practical Motivation} tasks (Tasks 3 and 4), since options for the \textit{Definition} tasks (Tasks 1 and 2) correspond directly to standard theoretical definitions and do not require generation. Specifically, we utilize a Motivation Analyst designed to generate practically grounded motivation options based on character behavior, story context, and visual cues. The analyst operates in two modes corresponding to distinct psychological frameworks: Maslow's Hierarchy of Needs and Reiss's 16 Basic Desires, referring to the definitions shown in Tables \ref{tab:maslow_definitions} and \ref{tab:reiss_definitions}.

Two distinct LLM-based analysts (Grok-4.1-Fast and Gemini-3) independently generate sets of practical motivation options based on the verified behavior chains. These models then participate in a cross-model review mechanism, providing feedback on: \begin{enumerate} \item Logical soundness of the options. \item Correctness of the answer based on visual and narrative context. \item Strict alignment with Maslow's and Reiss's definitions to minimize hidden biases from a single model. \end{enumerate}

The feedback is compiled by a Validator (Grok-4.1-Fast-reasoning) to select the highest-quality candidates and form the final option sets. The prompt templates for the Motivation Analyst, Reviewer, and Validator for both Maslow and Reiss tasks are presented in Tables \ref{tab:analyst_prompt_generic}, \ref{tab:validator_prompt}, and \ref{tab:validator_selector_prompt}.

\paragraph{Option-artifact controls}
\label{app:option_artifacts}
We audit whether surface properties correlate with correctness in the generated Practical options. Table~\ref{tab:option_artifacts} shows small mean length differences, modest associations between lead-in templates and correctness, and non-uniform gold-label frequencies. These descriptive controls do not rule out combinations of lexical cues, so option artifacts remain a limitation rather than a closed concern.

\begin{table}[t]
\centering
\small
\setlength{\tabcolsep}{3.5pt}
\begin{tabular}{lrrrr}
\toprule
Theory & $\Delta$ words & Cram\'{e}r's $V$ & Top label & Norm. $H$ \\
\midrule
Maslow & +0.78 & 0.182 & 36.28\% & 0.820 \\
Reiss  & +0.27 & 0.133 & 24.09\% & 0.854 \\
\bottomrule
\end{tabular}
\caption{Practical-option artifact controls. $\Delta$ words is the mean correct-minus-distractor word count; $V$ measures association between lead-in template and correctness; Top label is the most frequent gold label; and Norm. $H$ is normalized label entropy.}
\label{tab:option_artifacts}
\end{table}

\subsection{Details of the annotation process} \label{appendix_annotation}
The gold-label annotators were three computer-science graduate students with prior experience in Theory-of-Mind benchmark annotation. They received explanations of the categories in Tables~\ref{tab:maslow_definitions} and \ref{tab:reiss_definitions} from a researcher with a psychology background, practiced on representative examples, and received feedback on typical errors during a first-100-instance calibration phase. They subsequently annotated the retained items using the supplied story context, images, behavior description, and explicit theory definitions. The interface records one unordered multi-label Maslow set and one Reiss set for each behavior; the same set supplies the answer for the corresponding Definition and Practical formulations. Judgments were made independently and then aggregated. The three evaluators used for the human baseline were a different group and had no role in gold-label construction.

All gold-label annotators and human-baseline evaluators were informed that their responses and aggregate evaluation results would be used in this research and reported in an academic publication, and each provided explicit consent. We do not release identifying participant information.

\begin{figure*}[h]
    \centering
    \label{anno}
    \begin{subfigure}[b]{0.48\textwidth}
        \centering
        \includegraphics[width=\linewidth]{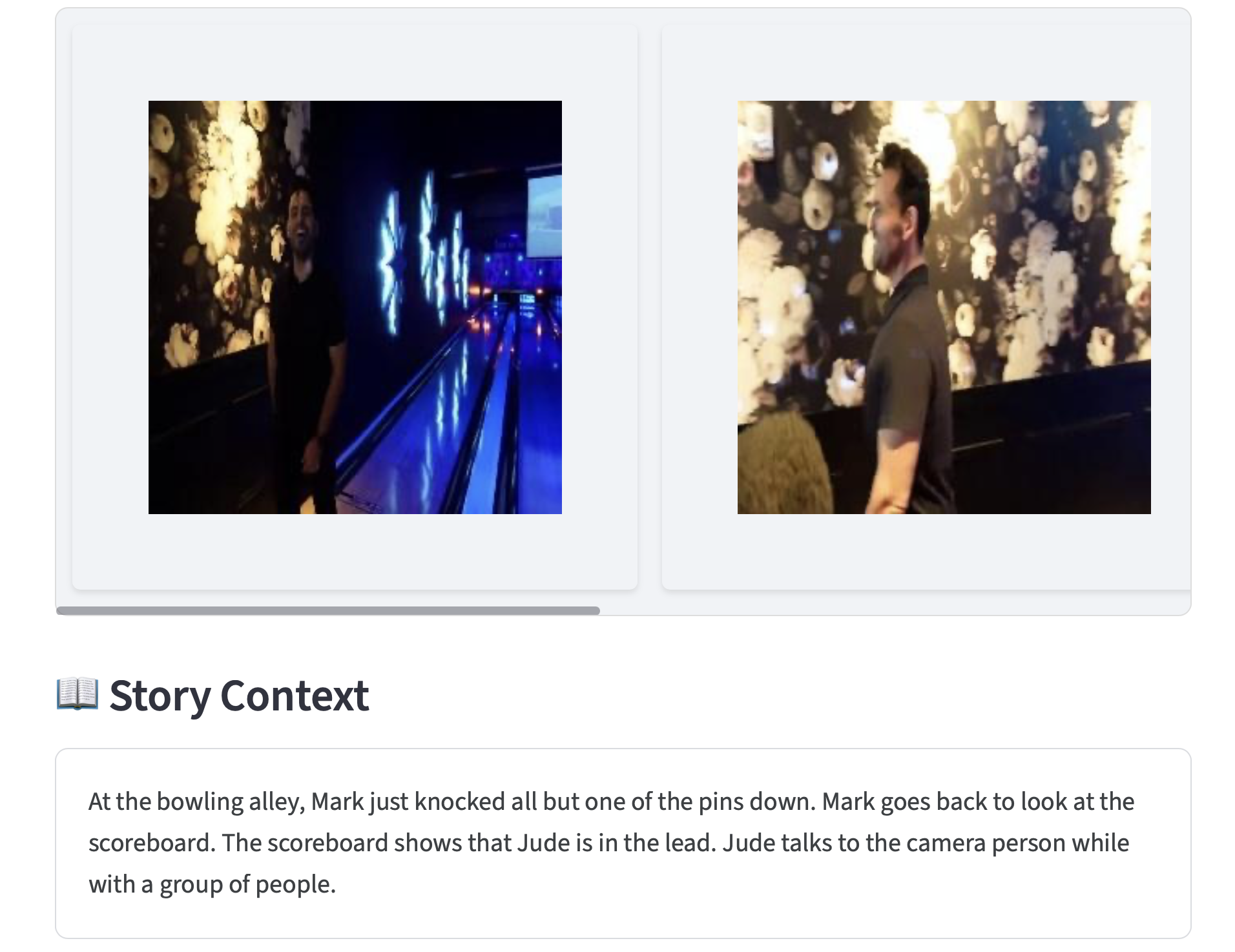}
        \caption{The initial step of the annotation framework.}
        \label{fig:anno1_step}
    \end{subfigure}
    \hfill
    \begin{subfigure}[b]{0.48\textwidth}
        \centering
        \includegraphics[width=\linewidth]{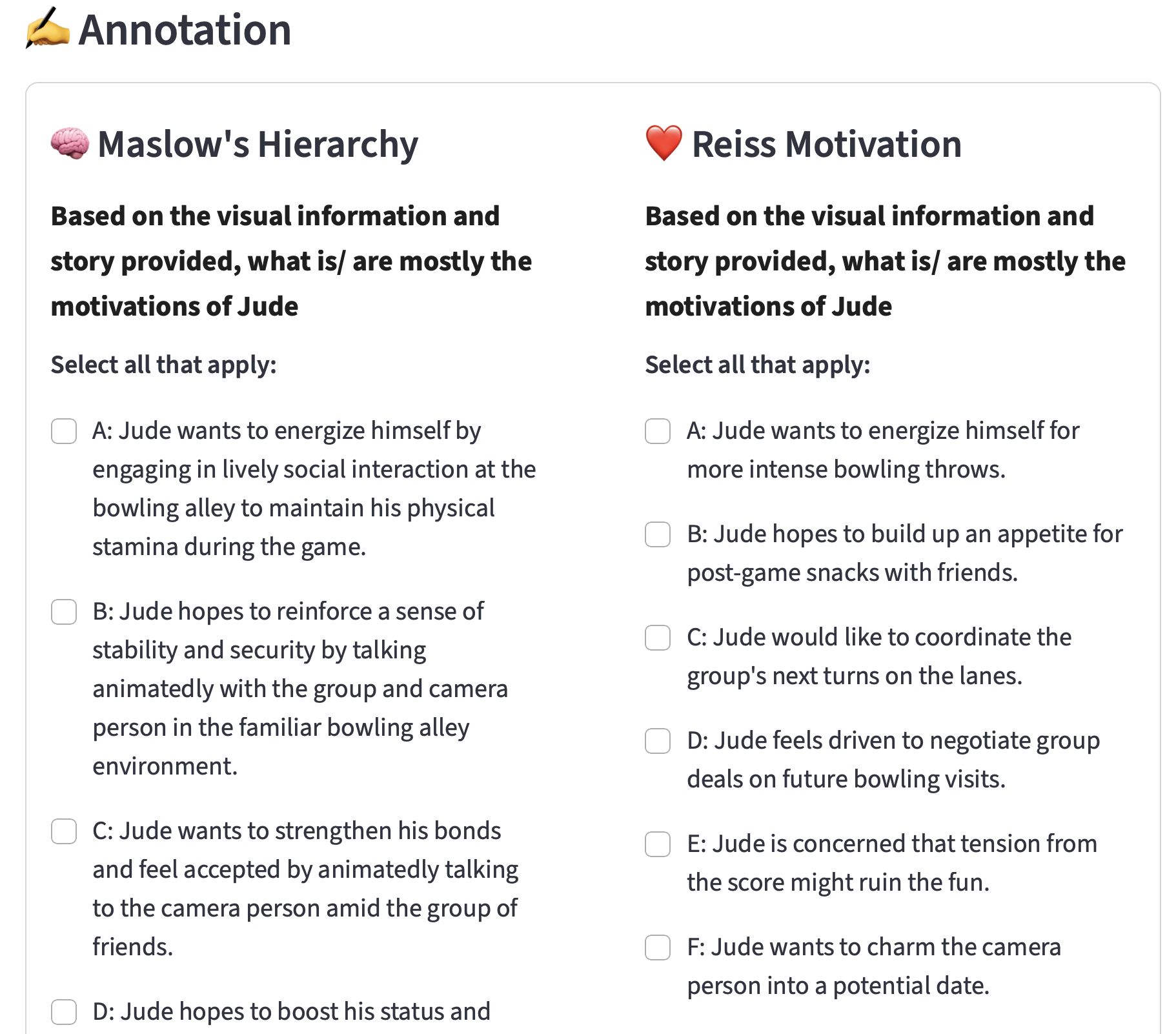}
        \caption{The validation step of the annotation framework.}
        \label{fig:anno2_step}
    \end{subfigure}

    \caption{Overview of the framework provided to the annotators. Panel (a) illustrates the interface for initial annotations, and panel (b) shows the review and validation interface.}
    \label{fig:annotation_framework_overview}
\end{figure*}

The annotation interface and instructions are illustrated in Figure~\ref{fig:annotation_framework_overview}.

\begin{table*}[t]
\centering
\footnotesize
\begin{tabularx}{\textwidth}{p{0.17\textwidth}p{0.20\textwidth}p{0.23\textwidth}X}
\toprule
Context & Target behavior & Gold labels & Why multi-label \\
\midrule
A cowboy rides with friends toward a forest while leading an extra horse. & Leads the group into the forest while holding the extra horse's reins. & Maslow: Love/Belonging, Esteem; Reiss: Social Contact, Power. & Companionship and a shared adventure support affiliation, while directing the group's movement and handling the extra horse support leadership and influence. Neither aspect alone fully captures the behavior. \\
\bottomrule
\end{tabularx}
\caption{Representative multi-label case. The explicit taxonomies allow multiple jointly supported interpretations rather than forcing a single gold category.}
\label{tab:multilabel_example}
\end{table*}

\subsection{Details of the Four Task Types}
\label{Appendix:question_type}
In this section, we describe the four question types used in \textsc{MulTivationBench}. For the Definition tasks, the question stem is: ``Based on the visual information and the story provided, which [theory label(s)] is/are most strongly expressed or fulfilled by the behavior of [character]?'' The options correspond to the standard theory definitions. For the Practical Motivation tasks, the question stem is: ``Based on the visual information and the story provided, what is/are the most likely motivation(s) behind [character]'s behavior?'' The options are context-specific motivations generated from the corresponding theory and the accumulated narrative context.

\subsection{Detailed Statistics of \textsc{Multivationbench}}
\label{Appendix:stats}
Table~\ref{tab:stats} presents story- and behavior-level statistics. Table~\ref{tab:label_cardinality_distribution} reports gold label-set sizes, and Table~\ref{tab:cardinality_performance} stratifies normal-mode performance by that size.

\begin{table}[t]
\centering
\small
\begin{tabular*}{\columnwidth}{@{\extracolsep{\fill}}lccccc}
\toprule
Theory & C1 & C2 & C3 & C4 & Multi. \\
\midrule
Maslow & 3,751 & 251 & 21 & -- & 272 / 6.76\% \\
Reiss & 3,645 & 317 & 59 & 2 & 378 / 9.40\% \\
\bottomrule
\end{tabular*}
\caption{Gold label-set cardinality distribution. C$k$ denotes behavior points with exactly $k$ labels. Multi. reports the number and percentage with more than one label.}
\label{tab:label_cardinality_distribution}
\end{table}

\begin{table}[t]
\centering
\small
\begin{tabular*}{\columnwidth}{@{\extracolsep{\fill}}lrrr}
\toprule
Theory & Gold card. & EM & F1 \\
\midrule
Maslow & 1 & 36.45 & 55.50 \\
       & 2 & 9.55  & 53.82 \\
       & 3 & 2.48  & 43.83 \\
Reiss  & 1 & 25.94 & 38.81 \\
       & 2 & 4.06  & 39.70 \\
       & 3 & 0.64  & 40.04 \\
       & 4 & 0.00  & 32.20 \\
\bottomrule
\end{tabular*}
\caption{Normal-mode EM/F1 (\%) stratified by gold cardinality, macro-averaged over retained models with Definition and Practical variants pooled. Reiss cardinality 4 contains only two behaviors and should not be overinterpreted.}
\label{tab:cardinality_performance}
\end{table}

EM falls much more sharply than F1 as the gold set grows. From cardinality 1 to 2, Maslow EM drops 26.90 points while F1 drops 1.68; Reiss EM drops 21.88 points while F1 increases by 0.89. This confirms that EM is especially stringent for multi-label cases, whereas example-based F1 preserves partial overlap.

\subsection{Release and Reproducibility}
\label{app:release_reproducibility}
Table~\ref{tab:release_artifacts} summarizes the release boundary and executable path. The release-facing \texttt{submission/gt.json} stores source identifiers, behavior annotations, contextual and standard-definition questions/options, and Maslow/Reiss answer sets. Evaluation prompts are expanded by \texttt{tools/real\_evalute.py}; \texttt{llm\_response/calculate\_new.py} computes the headline EM/F1 table, while \texttt{calculate\_metric\_v2.py} supports detailed analyses. The repository README gives commands and expected data layouts.

\begin{table*}[t]
\centering
\footnotesize
\begin{tabularx}{\textwidth}{p{0.18\textwidth}p{0.31\textwidth}p{0.23\textwidth}X}
\toprule
Source / artifact & Released material & Reconstruction & Terms / restriction \\
\midrule
\texttt{submission/gt.json} & Derived behaviors, labels, four task variants, source IDs & Directly usable after licensed context is restored & New annotations released within upstream constraints \\
MovieBench & No original videos or frames redistributed & Obtain source videos; extract frames from stored timestamps (not fully automated) & CC BY 4.0 \\
StoryReasoning & No original images or story text redistributed & Official loader plus the provided restore script & CC BY-ND 4.0; explicit permission for released annotations \\
SSID & No original images or story text redistributed & Official data plus the provided restore script & CC BY-NC-ND 4.0; non-commercial research only; explicit permission \\
Evaluation code & Prompt construction, parsing, EM/F1 scoring, analysis scripts & Run on reconstructed benchmark and model outputs & Repository documentation specifies entry points \\
\bottomrule
\end{tabularx}
\caption{Release contents, reconstruction workflow, and source-specific constraints. Original source media and story text are not redistributed.}
\label{tab:release_artifacts}
\end{table*}

\section{Appendix for Experiment }
\subsection{Prompts for the Four Task Types}
\label{Appendix:4prompt}
In this section, we detail the specific prompts used for the four distinct question configurations in our benchmark. Tables \ref{tab:maslow_definition_raw_prompt}, \ref{tab:maslow_practical_raw_prompt}, \ref{tab:reiss_definition_raw_prompt}, and \ref{tab:reiss_practical_raw_prompt} provide concrete examples of the data instances passed to the Multimodal Large Language Model (MLLM) to elicit the final choice selections.

These templates cover both the definition-based tasks (Tables \ref{tab:maslow_definition_raw_prompt} and \ref{tab:reiss_definition_raw_prompt}), which require the model to map behaviors directly to standard theoretical definitions, and the Practical Motivation tasks (Tables \ref{tab:maslow_practical_raw_prompt} and \ref{tab:reiss_practical_raw_prompt}), which utilize context-specific motivation options. Crucially, to mitigate position bias---where models may preferentially select options based on their index rather than content---the order of the multiple-choice options is randomized for every instance prior to inference.

\subsection{Prompt Variations for Different Modalities}
To evaluate the contribution of different modalities to the reasoning process, we adjust the prompt header and context block accordingly. Table \ref{tab:modality_prompts} details how the introduction string and story context are modified for the \textbf{Multimodal}, \textbf{Image-Only}, and \textbf{Text-Only} settings.
\subsection{Metric Details}
\label{app:metric-details}

\paragraph{Example-based F1}
In our setting, Exact Match (EM) is a strict set-level metric: it only counts a prediction as correct when the predicted option set exactly matches the gold set. While useful, this criterion does not distinguish between a completely wrong prediction and a partially correct one. We therefore also report \textit{Example-based F1}, which captures instance-level partial agreement between the predicted and gold option sets.

Let $G_i$ denote the gold option set and $\hat{G}_i$ denote the predicted option set for test instance $i$.
For each instance, we first compute precision and recall:
\[
P_i = \frac{|G_i \cap \hat{G}_i|}{|\hat{G}_i|},
\qquad
R_i = \frac{|G_i \cap \hat{G}_i|}{|G_i|}.
\]
The instance-level F1 score is then
\[
F1_i = \frac{2P_iR_i}{P_i + R_i}
= \frac{2|G_i \cap \hat{G}_i|}{|G_i| + |\hat{G}_i|}.
\]
The final example-based F1 is obtained by averaging over all $N$ test instances:
\[
\text{Example-based F1} = \frac{1}{N}\sum_{i=1}^{N} F1_i.
\]

This metric gives partial credit when the predicted and gold option sets overlap, while still penalizing both false positives and false negatives. In this way, it complements EM by distinguishing near-miss predictions from fully incorrect ones.

\subsection{Human Performance Evaluation}
\label{App:human_evaluation}

To measure human performance on \textsc{MulTivationBench}, we employ three graduate-student evaluators, distinct from the gold-label annotators, to complete the same benchmark tasks as the models. Each instance is evaluated under four settings: Maslow Definition, Maslow Practical Motivation, Reiss Definition, and Reiss Practical Motivation. The questions and instructions follow the four task types in Appendix~\ref{Appendix:question_type}. We compute the final human predictions by majority vote over the three evaluators' responses.
This results in 70.6\% EM and 81.5\% F1 on Maslow Definition, 78.6\% EM and 87.6\% F1 on Maslow Practical Motivation, 60.7\% EM and 74.2\% F1 on Reiss Definition, and 63.9\% EM and 72.9\% F1 on Reiss Practical Motivation.

\subsection{Common-Subset Comparability Check}
\label{App:sub_common}

Phi-3.5-Vision-instruct cannot process the Long-story multimodal and image-only subsets due to image context-window limitations. As a result, its Overall score in Table~\ref{tab:master_results} is computed only over the successfully processed subsets and is less directly comparable to models evaluated on all story lengths.

To check whether this missing subset changes the relative model comparison, we recompute normal-mode model performance on the exact intersection of task instances shared by all retained models. The common subset contains 14,180 task instances. As shown in Table~\ref{tab:common_subset}, all retained models preserve the same EM and F1 ranks under the common-subset evaluation. Phi-3.5-Vision-instruct obtains 29.10 EM / 36.21 F1 both before and after common-subset filtering, indicating that the missing Long-story subset does not change its relative position under this check. We therefore keep the original full evaluation in the main paper while explicitly noting the context-window caveat.

\begin{table}[t]
\centering
\small
\begin{tabular*}{\columnwidth}{@{\extracolsep{\fill}}lcc}
\toprule
Model & Full & Common \\
\midrule
Phi-4-Multimodal & 39.22 / 42.79 & 39.22 / 43.03 \\
Gemini-3-Flash & 35.72 / 53.73 & 36.09 / 54.22 \\
Llama-Scout & 32.60 / 48.97 & 33.48 / 49.60 \\
Grok-4.1-Fast & 31.77 / 54.80 & 32.53 / 55.32 \\
Llama-Maverick & 30.31 / 49.22 & 30.28 / 49.81 \\
Phi-3.5-Vision & 29.10 / 36.21 & 29.10 / 36.21 \\
o4-mini & 26.61 / 53.13 & 27.29 / 53.62 \\
Nemotron-Nano-12B & 8.22 / 38.08 & 8.58 / 38.16 \\
\bottomrule
\end{tabular*}
\caption{Common-subset comparability check. Each cell reports EM / F1. ``Full'' reports the original retained-instance score, while ``Common'' recomputes scores on the exact intersection of task instances shared by all retained models.}
\label{tab:common_subset}
\end{table}

We compute 95\% confidence intervals with 10{,}000 whole-story bootstrap samples (Table~\ref{tab:common_subset_ci}). Pairwise differences use 20{,}000 whole-story sign-flip permutations with Holm correction applied separately to the 28 EM and 28 F1 comparisons. In each metric, 26 of 28 pairs remain significant. The two leading differences are also reliable: Phi-4 exceeds Gemini-3-Flash by 3.13 EM points ($p_H=.0014$), and Grok-4.1-Fast exceeds Gemini-3-Flash by 1.10 F1 points ($p_H=.0042$). These tests support clearly separated scores without treating every small rank gap as meaningful.

\begin{table*}[t]
\centering
\small
\begin{tabular*}{\textwidth}{@{\extracolsep{\fill}}lcc}
\toprule
Model & Common EM [95\% CI] & Common F1 [95\% CI] \\
\midrule
Phi-4-Multimodal & 39.22 [38.00, 40.47] & 43.03 [41.86, 44.22] \\
Gemini-3-Flash & 36.09 [34.83, 37.38] & 54.22 [53.14, 55.31] \\
Llama-4-Scout & 33.48 [32.29, 34.72] & 49.60 [48.52, 50.71] \\
Grok-4.1-Fast & 32.53 [31.35, 33.75] & 55.32 [54.32, 56.35] \\
Llama-4-Maverick & 30.28 [29.14, 31.42] & 49.81 [48.79, 50.86] \\
Phi-3.5-Vision & 29.10 [28.02, 30.22] & 36.21 [35.17, 37.30] \\
o4-mini & 27.29 [26.17, 28.44] & 53.62 [52.63, 54.62] \\
Nemotron-Nano-12B & 8.58 [7.98, 9.20] & 38.16 [37.39, 38.94] \\
\bottomrule
\end{tabular*}
\caption{Normal-mode common-subset scores and whole-story bootstrap confidence intervals over 14{,}180 tasks, 3{,}545 behavior points, and 1000 stories.}
\label{tab:common_subset_ci}
\end{table*}

\subsection{Same-Behavior Context Ablation}
\label{App:same_behavior_ablation}

To more directly test whether model predictions for the same behavior change when accumulated context is altered, we conduct a same-behavior context-ablation analysis. Unlike the standard evaluation setting, where each behavior is evaluated with the accumulated multimodal context available up to its image index, this analysis keeps the target behavior fixed while removing earlier context units from the input.

Specifically, we sample 150 final behavior points from \textsc{MultivationBench}. For each sampled behavior, we preserve the same target character, target behavior, current image, and answer options, but remove one, two, or three earlier text-image units from the accumulated context. We then re-evaluate model predictions under these reduced-context settings. We report EM and example-based F1 averaged over the Maslow and Reiss Practical Motivation tasks.

\begin{table*}[t]
\centering
\small
\begin{tabular*}{\textwidth}{@{\extracolsep{\fill}}lccccc}
\toprule
Model & Full & Drop1 & Drop2 & Drop3 & $\Delta$F1 \\
\midrule
Grok-4.1-Fast & 21.33 / 52.51 & 22.00 / 50.20 & 22.00 / 50.76 & 22.67 / 49.07 & -3.44 \\
Llama-4-Scout & 39.67 / 48.75 & 41.00 / 49.14 & 41.33 / 49.61 & 38.67 / 48.48 & -0.27 \\
Llama-4-Maverick & 21.00 / 45.97 & 27.33 / 45.88 & 26.33 / 47.82 & 20.33 / 44.68 & -1.28 \\
Phi-4-multimodal & 33.67 / 38.26 & 33.00 / 37.77 & 33.33 / 37.55 & 36.33 / 39.21 & +0.94 \\
Nemotron-Nano-12B & 1.33 / 38.62 & 1.33 / 38.92 & 1.33 / 38.47 & 1.00 / 38.52 & -0.10 \\
\midrule
Average & 23.40 / 44.82 & 24.93 / 44.38 & 24.87 / 44.84 & 23.80 / 43.99 & -0.83 \\
\bottomrule
\end{tabular*}
\caption{Same-behavior context-ablation results. Each cell in the Full, Drop1, Drop2, and Drop3 columns reports EM / F1, averaged over the Maslow and Reiss Practical Motivation tasks. Full uses the complete accumulated context for the sampled behavior. Drop1, Drop2, and Drop3 remove one, two, and three earlier text-image context units, respectively, while keeping the target character, target behavior, current image, and answer options fixed. $\Delta$F1 denotes Drop3 F1 minus Full F1.}
\label{tab:same_behavior_ablation}
\end{table*}

As shown in Table~\ref{tab:same_behavior_ablation}, removing earlier context does not produce a uniform degradation across models. Average F1 changes from 44.82 under full context to 44.38, 44.84, and 43.99 after dropping one, two, and three earlier text-image units, respectively. This suggests that current MLLMs do not consistently exploit distant earlier context when predicting the motivation of a fixed later behavior. Some models show stronger context sensitivity, such as Grok-4.1-Fast, whose F1 decreases by 3.44 points after dropping three earlier units. Overall, this analysis indicates that motivation revision under changing accumulated context remains limited and model-dependent.

\subsection{Softer Story-Level Metrics}
\label{App:soft_story_metrics}

The full-story consistency metric in Table~\ref{tab:story_consistency} is intentionally strict: a story is counted as correct only if every question is answered with exact-match correctness. This provides a stress-test of whether models can maintain fully correct motivation reasoning across an entire narrative. However, because this all-correct criterion can compress differences among models, we additionally report softer story-level metrics.

For each model and story, we first average EM and example-based F1 over all questions within that story. We then macro-average these per-story scores across all stories. We refer to these metrics as macro story EM and macro story F1. Unlike full-story consistency, these metrics preserve partial story-level performance while still evaluating models at the story level rather than only at the individual-question level.

\begin{table}[t]
\centering
\small
\begin{tabular*}{\columnwidth}{@{\extracolsep{\fill}}lccc}
\toprule
Model & Full & Story EM & Story F1 \\
\midrule
Phi-4-Multimodal & 0.80 & 38.51 & 42.61 \\
Gemini-3-Flash & 0.50 & 35.48 & 53.74 \\
Llama-Scout & 0.30 & 32.94 & 49.03 \\
Grok-4.1-Fast & 0.20 & 32.05 & 54.80 \\
Llama-Maverick & 0.10 & 29.86 & 49.02 \\
Phi-3.5-Vision & 0.10 & 28.75 & 36.21 \\
o4-mini & 0.10 & 27.14 & 53.18 \\
Nemotron-Nano-12B & 0.00 & 8.65 & 38.22 \\
\bottomrule
\end{tabular*}
\caption{Softer story-level metrics (\%). Full is the all-correct story consistency score, where a story is counted as correct only if every question is answered with exact-match correctness. Story EM and Story F1 first average EM/F1 over questions within each story and then macro-average across stories.}
\label{tab:soft_story_metrics}
\end{table}

As shown in Table~\ref{tab:soft_story_metrics}, the all-correct full-story consistency score ranges only from 0.00\% to 0.80\%, showing that perfectly correct whole-story performance remains rare for current MLLMs. In contrast, macro story EM ranges from 8.65\% to 38.51\%, and macro story F1 ranges from 36.21\% to 54.80\%, revealing clearer differences among models. These softer metrics therefore complement the strict consistency score: full-story consistency measures whether a model can answer an entire narrative perfectly, while macro story EM/F1 provide a more graded view of story-level performance.

\subsection{Paired Modality Tests}
\label{app:paired_modality}

Table~\ref{tab:paired_modality} pairs each model's multimodal prediction with the same task record in the caption-enriched Text-Only and Image-Only settings. We resample whole stories for 20{,}000 bootstrap confidence intervals and use 100{,}000 story-cluster sign flips, correcting all 32 hypotheses together with Holm's method. This preserves each story's correlated questions and prevents modality differences from being attributed to different evaluated examples.

\begin{table*}[t]
\centering
\scriptsize
\resizebox{\textwidth}{!}{\begin{tabular}{lcccc}
\toprule
& \multicolumn{2}{c}{Multimodal $-$ Text-Only} & \multicolumn{2}{c}{Multimodal $-$ Image-Only} \\
\cmidrule(lr){2-3}\cmidrule(lr){4-5}
Model & $\Delta$EM [95\% CI] & $\Delta$F1 [95\% CI] & $\Delta$EM [95\% CI] & $\Delta$F1 [95\% CI] \\
\midrule
Grok-4.1-Fast & $+1.14^{*}$ [0.64, 1.63] & $+0.81^{*}$ [0.44, 1.20] & $+8.53^{*}$ [7.40, 9.66] & $+17.77^{*}$ [16.67, 18.88] \\
Gemini-3-Flash & $+22.09^{*}$ [21.13, 23.04] & $+4.57^{*}$ [3.97, 5.15] & $+8.65^{*}$ [7.52, 9.78] & $+15.49^{*}$ [14.42, 16.54] \\
o4-mini & $+3.16^{*}$ [2.59, 3.73] & $+0.95^{*}$ [0.56, 1.34] & $+9.40^{*}$ [8.34, 10.44] & $+16.33^{*}$ [15.32, 17.34] \\
Llama-4-Maverick & $+21.80^{*}$ [20.85, 22.75] & $+2.70^{*}$ [2.11, 3.29] & $+5.47^{*}$ [4.38, 6.56] & $+15.71^{*}$ [14.68, 16.75] \\
Llama-4-Scout & $+5.95^{*}$ [5.31, 6.59] & $+0.68^{*}$ [0.23, 1.12] & $+4.16^{*}$ [3.05, 5.27] & $+14.84^{*}$ [13.78, 15.92] \\
Phi-4-Multimodal & $+33.32^{*}$ [32.14, 34.51] & $+5.94^{*}$ [5.06, 6.82] & $+12.99^{*}$ [11.99, 14.00] & $+14.30^{*}$ [13.30, 15.31] \\
Phi-3.5-Vision & $+11.39^{*}$ [10.50, 12.28] & $-2.21^{*}$ [-2.89, -1.54] & $+18.31^{*}$ [17.37, 19.26] & $+13.07^{*}$ [12.20, 13.95] \\
Nemotron-Nano-12B & $-0.65^{*}$ [-1.15, -0.16] & $-4.10^{*}$ [-4.62, -3.60] & $-0.01$ [-0.66, 0.64] & $+6.79^{*}$ [6.00, 7.56] \\
\bottomrule
\end{tabular}}
\caption{Paired modality effects in percentage points. Positive values favor multimodal input. $^{*}$ denotes $p_H<.05$ after a global Holm correction across all 32 model--comparison--metric hypotheses. Phi-3.5 uses its 14{,}180-task intersection; other coverage differences follow each model's documented image limit.}
\label{tab:paired_modality}
\end{table*}

Overall, 31 of 32 effects are significant: 28 favor multimodal input and three favor Text-Only. The single nonsignificant effect is Nemotron's multimodal--Image-Only EM difference. The direction and magnitude are therefore model-dependent, despite a consistent multimodal F1 advantage over Image-Only.

\subsection{Adjacent Behavior-Transition Analysis}
\label{app:adjacent_transitions}

We identify 3{,}023 consecutive behavior pairs across 974 stories. Gold Maslow sets change in 51.94\% of pairs, Reiss sets change in 63.45\%, and at least one theory changes in 70.49\% (2{,}131 pairs); 892 pairs retain both signatures. Among the 3{,}020 transitions for which accumulated image context strictly increases, the last rate is 70.53\%. Thus, most local transitions require the model to update at least part of the target label set.

For a stringent local diagnostic, a pair passes only when all four task components are exactly correct at both adjacent behaviors. Table~\ref{tab:transition_results} reports this pass rate and the mean F1 across the eight component predictions in each pair. Gold-change pairs are consistently harder than gold-stay pairs, and even the best strict pair pass is 1.65\%.

\begin{table}[t]
\centering
\scriptsize
\setlength{\tabcolsep}{3pt}
\begin{tabular*}{\columnwidth}{@{\extracolsep{\fill}}lrrrr}
\toprule
Model & All & Change & Stay & Pair F1 \\
\midrule
Phi-4-Multimodal & 1.65 & 0.05 & 5.49 & 42.63 \\
Gemini-3-Flash & 1.59 & 0.05 & 5.27 & 53.62 \\
Grok-4.1-Fast & 1.09 & 0.00 & 3.70 & 54.51 \\
Llama-4-Scout & 0.96 & 0.00 & 3.25 & 48.57 \\
Llama-4-Maverick & 0.46 & 0.00 & 1.57 & 49.23 \\
Phi-3.5-Vision & 0.40 & 0.00 & 1.35 & 31.78 \\
o4-mini & 0.40 & 0.05 & 1.23 & 52.84 \\
Nemotron-Nano-12B & 0.07 & 0.00 & 0.22 & 35.71 \\
\bottomrule
\end{tabular*}
\caption{Adjacent-pair results (\%). All, Change, and Stay are strict pair-pass rates overall and by whether either gold theory changes. Pair F1 averages component-level F1 across both behaviors and all four task types.}
\label{tab:transition_results}
\end{table}

\begin{algorithm*}[t]
\caption{Automated Multi-Stage Option Generation with Cross-Review}
\label{alg:option_gen}
\begin{algorithmic}[1]
\Require Behavior Chain $\mathcal{B} = \{Char, Behaviors[], ImgIndices[]\}$, Story $\mathcal{S}$
\Ensure Set of Validated Options $\mathcal{O}_{final}$

\State $\mathcal{O}_{final} \gets \emptyset$

\For{$k \gets 1$ to length($Behaviors$)}
    \State $b_k \gets Behaviors[k]$
    \State $idx_k \gets ImgIndices[k]$
    
    \State \textbf{Step 1: Context Accumulation}
    \State $\mathcal{C}_k \gets$ \Call{GetContext}{$\mathcal{S}$, $0 \dots idx_k$} 
    \Comment{Accumulate text \& images up to current event}
    
    \State $Valid \gets \text{False}$
    
    \While{\textbf{not} $Valid$}
        \State \textbf{Step 2: Independent Generation}
        \State $Op_X \gets$ \Call{ModelX.Generate}{$Char, b_k, \mathcal{C}_k$}
        \State $Op_Y \gets$ \Call{ModelY.Generate}{$Char, b_k, \mathcal{C}_k$}
        
        \State \textbf{Step 3: Cross-Model Review}
        \State $Feed_{X \to Y} \gets$ \Call{ModelX.Review}{$Op_Y, \mathcal{C}_k$}
        \State $Feed_{Y \to X} \gets$ \Call{ModelY.Review}{$Op_X, \mathcal{C}_k$}
        
        \State \textbf{Step 4: Validator Adjudication}
        \State $Selection \gets$ \Call{AgentZ.Validate}{$Op_X, Op_Y, Feed_{X \to Y}, Feed_{Y \to X}$}
        
        \If{$Selection \neq \text{NULL}$}
            \State $\mathcal{O}_{final}$.append($Selection$)
            \State $Valid \gets \text{True}$
        \Else
            \State \textbf{continue} \Comment{Validation failed; Re-run generation}
        \EndIf
    \EndWhile
\EndFor

\State \Return $\mathcal{O}_{final}$
\end{algorithmic}
\end{algorithm*}

\begin{table*}[h]
    \centering
    \small
    \renewcommand{\arraystretch}{1.2}
    \begin{tabular}{|p{0.95\linewidth}|}
        \hline
        \rowcolor{gray!15} \textbf{\large Extractor Prompt Template} \\
        \hline
        \vspace{2pt}
        \textbf{Role:} You are an Experienced Story Annotator and Narrative Architect. \\
        \textbf{Goal:} Analyze the story text and images to extract the Behavior Chain for main characters. \\
        \vspace{2pt}
        \textbf{TASKS:}
        \begin{enumerate}[leftmargin=1.5em, nosep, label=\arabic*.]
            \item \textbf{Plot Integrity:} Verify that selected characters appear in at least 2 frames.
            \item \textbf{Visual Validation:} Ensure textual descriptions align with visual details in the images.
            \item \textbf{Character Selection:} Select max 3 main characters based on narrative frequency and visual significance. Prioritize the top 2 if many appear.
            \item \textbf{Behavior Grouping:} Treat groups performing identical actions as a single character unit.
            \item \textbf{Conciseness:} Provide brief, context-aligned behavior descriptions per frame.
            \item \textbf{Presence Check:} Exclude characters who do not appear in the current frame.
            \item \textbf{Accuracy:} Cross-reference extracted behaviors with the original story text.
            \item \textbf{Name Consistency:} Map generic names (e.g., ``The man'') to specific names found in the text (e.g., ``John'').
            \item \textbf{Textual Grounding:} Exclude characters present in images but never mentioned in the text.
            \item \textbf{Implied Context:} Refine context for transition frames where human behavior is implied by the plot but not explicitly visible.
            \item \textbf{Motivation Generation}: Analyze the motivation of each proactive ,visible behavior made by the major character
            \item \textbf{Motivation Filter:} Remove characters that lack motivated behaviors throughout the story.
            \item \textbf{Observability:} Record only proactive, visible actions/expressions. Do NOT infer internal states.
            \item \textbf{De-duplication:} For continuous behaviors across multiple frames, record the behavior \textit{only} in the starting frame (e.g., only log ``sit'' in frame 1, not 2--4).
        \end{enumerate}
        \vspace{5pt}
        \textbf{Output:} JSON string only. \\
        \hline
    \end{tabular}
    \caption{The Extractor Prompt used in data preprocessing to identify main characters and validate behaviors.}
    \label{tab:extractor_prompt}
\end{table*}

\begin{table*}[h]
    \centering
    \small
    \renewcommand{\arraystretch}{1.2}
    \begin{tabular}{|p{0.95\linewidth}|}
        \hline
        \rowcolor{gray!15} \textbf{\large Reviewer Prompt Template} \\
        \hline
        \vspace{2pt}
        \textbf{Role:} You are an Expert Psychologist and Narrative Reviewer. \\
        \textbf{Goal:} Evaluate the extracted behavior chains to select the most accurate and grounded analysis. \\
        \vspace{2pt}
        \textbf{TASKS:}
        \begin{enumerate}[leftmargin=1.5em, nosep, label=\arabic*.]
            \item \textbf{Input Analysis:} Review the provided images, full story text, and the candidate behavior chains extracted by previous agents.
            \item \textbf{Significance Check:} Identify which characters are the most active and appear in the majority of the frames. Focus verification on these key figures.
            \item \textbf{Grounding Verification:} Rigorously check if the reported behaviors and motivations are supported by specific visual cues and the narrative context.
            \item \textbf{Validation Logic:} 
            \begin{itemize}[leftmargin=1em, label=-]
                \item If a behavior or motivation is hallucinated, inaccurate, or unsupported by the story, mark the chain as \textbf{Invalid} and provide specific reasoning.
                \item If the chain is accurate, fully grounded in both visual cues and text, and correctly explains the character's psychology, mark it as \textbf{Valid}.
            \end{itemize}
            \item \textbf{Final Output:} Return the single best validated behavior chain that accurately reflects the main character's arc.
        \end{enumerate}
        \vspace{5pt}
        \textbf{Output:} JSON string with validation status and the selected chain. \\
        \hline
    \end{tabular}
    \caption{The Reviewer Prompt used to validate the accuracy of extracted behavior chains against visual and textual evidence.}
    \label{tab:reviewer_prompt}
\end{table*}

\begin{table*}[t]
    \centering
    \small
    \renewcommand{\arraystretch}{1.3}
    
    \begin{tabularx}{\linewidth}{|>{\bfseries}l|X|}
        \hline
        \rowcolor{gray!15} \textbf{Needs} & \textbf{Definition} \\
        \hline
        Physiological needs & Air, food, drink, shelter, warmth, sex, sleep, etc. \\
        \hline
        Safety needs & Protection from elements, security, order, law, stability, freedom from fear. \\
        \hline
        Love and belongingness needs & Friendship, intimacy, trust, and acceptance, receiving and giving affection and love. Affiliating, being part of a group (family, friends, work). \\
        \hline
        Esteem needs & Which Maslow classified into two categories: (i) esteem for oneself (dignity, achievement, mastery, independence) and (ii) the need to be accepted and valued by others (e.g., status, prestige). \\
        \hline
        Cognitive needs & Knowledge and understanding, curiosity, exploration, need for meaning and predictability. \\
        \hline
        Aesthetic needs & Appreciation and search for beauty, balance, form, etc. \\
        \hline
        Self-actualization needs & Realizing personal potential, self-fulfillment, seeking personal growth, and peak experiences. \\
        \hline
        Transcendence needs & A person is motivated by values that transcend beyond the personal self. Beyond self-actualization, they represent the human desire to connect with a higher reality, purpose, or the universe. This level emphasizes altruism, spiritual connection, and helping others achieve their potential. \\
        \hline
    \end{tabularx}
    \caption{Maslow's 8-Level Hierarchy of Needs}
    \label{tab:maslow_definitions}
\end{table*}

\begin{table*}[t]
    \centering
    \small
    \renewcommand{\arraystretch}{1.2}

    \begin{tabularx}{\linewidth}{|>{\bfseries}l|X|}
        \hline
        \rowcolor{gray!15} \textbf{Desire} & \textbf{Definition} \\
        \hline
        Physical Exercise & Drive for physical activity and movement. \\
        \hline
        Eating & Desire to eat. \\
        \hline
        Order & Desire to organize (including desire for ritual). \\
        \hline
        Saving & Desire to collect, value of frugality. \\
        \hline
        Tranquility & Desire to avoid anxiety, fear. \\
        \hline
        Romance & Desire for sex (including courting). \\
        \hline
        Family & Desire to raise own children. \\
        \hline
        Acceptance & Desire for approval. \\
        \hline
        Social Contact & Desire for peer companionship (desire to play). \\
        \hline
        Independence & Desire to be autonomous. \\
        \hline
        Vengeance & Desire to get even (including desire to compete, to win). \\
        \hline
        Honor & Desire to obey a traditional moral code. \\
        \hline
        Power & Desire to influence (including leadership; related to mastery). \\
        \hline
        Status & Desire for social standing. \\
        \hline
        Curiosity & Desire for knowledge. \\
        \hline
        Idealism & Desire to improve society (including altruism, justice). \\
        \hline
    \end{tabularx}
    \caption{Reiss's 16 Basic Desires}
    \label{tab:reiss_definitions}
\end{table*}

\begin{table*}[h]
    \centering
    \small
    \renewcommand{\arraystretch}{1.2}
    \begin{tabular}{|p{0.95\linewidth}|}
        \hline
        \rowcolor{gray!20} \textbf{\large Motivation Analyst Prompt Template} \\
        \hline
        \vspace{2pt}
        \textbf{Role:} You are a Professional Psychologist and Expert in [Theory Name]. \\
        \textbf{Goal:} Create high-quality, multimodal multiple-choice options based on the provided visual story. \\
        \vspace{2pt}
        \textbf{INPUTS:}
        \begin{itemize}[leftmargin=1.5em, nosep, label=-]
            \item \textbf{Story Background:} \{story\_context\}
            \item \textbf{Context Images:} \{context\_images\}
            \item \textbf{Current Scenario:} Character: \{target\_character\}, Behavior: \{behavior\}, Image: \{current\_image\}
        \end{itemize}
        \vspace{5pt}
        \textbf{THEORETICAL DEFINITIONS:} \\
        Use the following definitions ONLY as internal reference (Never include the full definition text in the output): \\
        \textbf{\{definitions\}}
        
        \vspace{5pt}
        \textbf{TASK:}
        \begin{enumerate}[leftmargin=1.5em, nosep, label=\arabic*.]
            \item Fully step into the role of \{target\_character\} and analyze their specific behavior in the context of the story.
            \item Create exactly [8 or 16] options. Each option must explain the practical motivation of the character corresponding to one level of the provided theory.
        \end{enumerate}

        \vspace{5pt}
        \textbf{CRITICAL GENERATION INSTRUCTIONS:}
        \begin{enumerate}[leftmargin=1.5em, nosep, label=\arabic*.]
            \item \textbf{Pattern:} Every option must be a single sentence: ``(\textit{theory\_label}) \{target\_character\} \textit{desire\_phrase} [practical goal supported by story actions].''
            \item \textbf{Phrasing:} Use desire phrases like ``wants'', ``hopes to'', ``feels driven'' to connect the character to the goal naturally.
            \item \textbf{Focus:} Focus ONLY on the motivation behind the SPECIFIC behavior listed above. Do not restate the behavior itself; explain the \textit{why}.
            \item \textbf{Constraint:} Do NOT include the full definition text in the output options.
        \end{enumerate}
        \vspace{5pt}
        \textbf{Question Wording:} ``Based on the visual information and the story provided, what is the most likely motivation of \{target\_character\}?'' \\
        \textbf{Output:} JSON string only. \\
        \hline
    \end{tabular}
    \caption{The generic Motivation Analyst Prompt. The \{definitions\} placeholder is dynamically populated with either Maslow's Hierarchy or Reiss's Basic Desires during generation.}
    \label{tab:analyst_prompt_generic}
\end{table*}

\begin{table*}[h]
    \centering
    \small
    \renewcommand{\arraystretch}{1.2}
    \begin{tabular}{|p{0.95\linewidth}|}
        \hline
        \rowcolor{gray!15} \textbf{\large Option reviewer Prompt Template} \\
        \hline
        \vspace{2pt}
        \textbf{Role:} You are a strict Quality Assurance Auditor. \\
        \textbf{Goal:} Review the provided multiple-choice options for specific errors and style violations. \\
        \vspace{2pt}
        \textbf{INPUTS:}
        \begin{itemize}[leftmargin=1.5em, nosep, label=-]
            \item \textbf{Story Text:} \{full\_text\}
            \item \textbf{Current Questions:} \{question\_set\_json\}
        \end{itemize}
        \vspace{5pt}
        \textbf{TASKS (CHECK FOR FAILURE):}
        \begin{enumerate}[leftmargin=1.5em, nosep, label=\arabic*.]
            \item \textbf{CHECK FOR HALLUCINATED NAMES (CRITICAL):}
            \begin{itemize}[leftmargin=1em, label=-]
                \item Scan every option for proper names (e.g., ``Sarah'', ``John'').
                \item Check if that name exists in the \textbf{STORY TEXT}.
                \item \textbf{FAIL} if a name appears in an option but NOT in the story text.
            \end{itemize}
            
            \item \textbf{CHECK FOR BEHAVIOR VS MOTIVATION:}
            \begin{itemize}[leftmargin=1em, label=-]
                \item Options must describe \textbf{INTERNAL MOTIVATION} (desires, hopes, goals).
                \item \textbf{FAIL} if an option describes \textbf{EXTERNAL BEHAVIOR} (actions, movements).
                \item \textit{Fail Example:} ``Simon engages customers energetically... because he wants to...'' (Contains action).
                \item \textit{Pass Example:} ``Simon wants to engage customers energetically for...'' (Pure motivation).
            \end{itemize}
            
            \item \textbf{CHECK LOGIC MAPPING:}
            \begin{itemize}[leftmargin=1em, label=-]
                \item Verify Maslow Option A is Physiological, B is Safety, etc.
                \item Verify Reiss Option A is Exercise, B is Eating, etc.
            \end{itemize}
        \end{enumerate}
        
        \vspace{5pt}
        \textbf{OUTPUT FORMAT:} \\
        If all checks pass: \texttt{\{ "status": "PASS" \}} \\
        If ANY check fails: \texttt{\{ "status": "FAIL", "feedback": "1. [Name Error]... 2. [Style Error]..." \}} \\
        \hline
    \end{tabular}
    \caption{The Validator Prompt used to rigorously audit generated questions for hallucinations, formatting errors, and logical consistency.}
    \label{tab:validator_prompt}
\end{table*}

\begin{table*}[h]
    \centering
    \small
    \renewcommand{\arraystretch}{1.2}
    \begin{tabular}{|p{0.95\linewidth}|}
        \hline
        \rowcolor{gray!15} \textbf{\large Option Validator \& Selector Prompt Template} \\
        \hline
        \vspace{2pt}
        \textbf{Role:} You are a Specialized Linguistic Editor and Lead Psychologist. \\
        \textbf{Goal:} Compare two sets of generated multiple-choice options and select the highest quality set based on strict psychological and linguistic standards. \\
        \vspace{2pt}
        \textbf{INPUTS:}
        \begin{itemize}[leftmargin=1.5em, nosep, label=-]
            \item \textbf{Story Text:} \{full\_text\}
            \item \textbf{Visual Context:} \{img\_count\} images attached.
            \item \textbf{Option Set A:} \{option\_set\_a\_json\}
            \item \textbf{Option Set B:} \{option\_set\_b\_json\}
        \end{itemize}
        \vspace{5pt}
        \textbf{EVALUATION CRITERIA (The "Best Set" Checklist):}
        \begin{enumerate}[leftmargin=1.5em, nosep, label=\arabic*.]
            \item \textbf{Hallucination Check (Critical):} 
            \begin{itemize}[leftmargin=1em, label=-]
                \item Does the set use names not present in the story? (e.g., using "John" when the text only says "The man").
                \item \textit{Penalty:} Immediate disqualification for hallucinated names.
            \end{itemize}
            
            \item \textbf{Motivation vs. Behavior:}
            \begin{itemize}[leftmargin=1em, label=-]
                \item Options must answer \textbf{WHY} (internal drive), not \textbf{WHAT} (external action).
                \item \textit{Bad:} ``(Social) He is playing tennis with friends.'' (Action description).
                \item \textit{Good:} ``(Social) He values the companionship found in shared activities.'' (Internal motivation).
                \item \textit{Preference:} Select the set that consistently focuses on internal psychological states.
            \end{itemize}
            
            \item \textbf{Sentence Variety (The "Anti-Robot" Rule):}
            \begin{itemize}[leftmargin=1em, label=-]
                \item Does the set sound repetitive (e.g., every option starts with ``He wants to...'')?
                \item \textit{Preference:} Select the set that uses varied patterns (e.g., ``She fears...'', ``He feels driven to...'', ``They prioritize...'', ``She envisions...'').
            \end{itemize}
        \end{enumerate}
        
        \vspace{5pt}
        \textbf{TASK:}
        \begin{enumerate}[leftmargin=1.5em, nosep, label=\arabic*.]
            \item Rigorously compare Set A and Set B against the criteria above.
            \item Select the superior set.
            \item If the winner still contains minor errors (e.g., one repetitive sentence), you may apply final edits to perfect it.
        \end{enumerate}
        
        \vspace{5pt}
        \textbf{OUTPUT FORMAT:} \\
        Return ONLY the final, polished JSON of the selected set. \\
        \hline
    \end{tabular}
    \caption{The Validator Prompt used to compare two candidate option sets and select the one with higher linguistic variety and psychological accuracy.}
    \label{tab:validator_selector_prompt}
\end{table*}

\begin{table*}[t]
\centering
\renewcommand{\arraystretch}{1.2}
\setlength{\tabcolsep}{10pt}
\begin{tabular}{l c ccc cc}
\toprule
\multirow{2}{*}{\textbf{Category} (Images)} & \multirow{2}{*}{\textbf{Stories}} & \multicolumn{3}{c}{\textbf{Complexity Metrics (Avg)}} & \multicolumn{2}{c}{\textbf{Total Annotations}} \\ 
\cmidrule(lr){3-5} \cmidrule(lr){6-7}
 &  & \textbf{Behav.} & \textbf{Img.} & \textbf{Ctx Tok.} & \textbf{Total Behav.} & \textbf{Total Qs} \\ 
\midrule
\textbf{Short} ($\leq 5$) & 411 & 3.01 & 4.03  & 87.8  & 1,236 & 4,944 \\
\textbf{Medium} (6--9)    & 384 & 4.15 & 7.21  & 384.7 & 1,593 & 6,372 \\
\textbf{Long} (10+)       & 205 & 5.82 & 12.57 & 560.4 & 1,194 & 4,776 \\
\midrule
\textbf{Total}    & \textbf{1,000} & \textbf{4.02} & \textbf{7.00} & \textbf{345.6} & \textbf{4,023} & \textbf{16,092} \\
\bottomrule
\end{tabular}
\caption{\textbf{Data statistics of our proposed \textsc{MultivationBench}.} (Behav.: average behaviors per story, Img.: average image count per story, Ctx Tok.: average context tokens, Total Behav.: total annotated behaviors, and Total Qs: total questions.)}
\label{tab:stats}
\end{table*}

\begin{table*}[h]
    \centering
    \small
    \renewcommand{\arraystretch}{1.2}
    \begin{tabular}{|p{0.95\linewidth}|}
        \hline
        \rowcolor{gray!15} \textbf{\large Maslow Practical Visual Reasoning Prompt} \\
        \hline
        \vspace{2pt}
        The following is a Visual Reasoning Multiple-Choice Question. Carefully read the story and view the images, fully immerse yourself in the role of the character described, and reason based on the information provided. Your answer should rely strictly on the given details. \\
        \vspace{5pt}
        \textbf{Story Context:} The tattooed woman with pink mohawk is excited and ready for her outdoor trip with her partner. The tattooed woman with pink mohawk is excitedly describing their outdoor trip plans. The tattooed woman with pink mohawk is riding in the vehicle, excited for the road trip. \\
        \vspace{5pt}
        \textbf{Question:} \\
        Based on the visual information and story provided, what is/ are mostly the motivation(s) of Tattooed woman with pink mohawk's behavior? \\
        \vspace{5pt}
        \textbf{Options:}
        \begin{itemize}[leftmargin=1.5em, nosep, label={}]
            \item \textbf{A.} Tattooed woman with pink mohawk wants to enjoy the road trip to satisfy her immediate needs for food, rest, or physical comfort during the outdoor journey.
            \item \textbf{B.} Tattooed woman with pink mohawk hopes to ensure a secure and stable road trip environment free from potential hazards while traveling.
            \item \textbf{C.} Tattooed woman with pink mohawk feels driven to enjoy the road trip to deepen her connection and share excitement with her partner.
            \item \textbf{D.} Tattooed woman with pink mohawk would like to embrace the road trip adventure to affirm her independence and achieve a sense of personal accomplishment.
            \item \textbf{E.} Tattooed woman with pink mohawk is eager to explore new sights during the road trip to satisfy her curiosity about the passing landscapes and destinations.
            \item \textbf{F.} Tattooed woman with pink mohawk wants to appreciate the scenic beauty unfolding outside the vehicle window during the road trip.
            \item \textbf{G.} Tattooed woman with pink mohawk hopes to embrace the road trip as a step toward personal growth through spontaneous adventure and self-discovery.
            \item \textbf{H.} Tattooed woman with pink mohawk feels driven to experience the road trip to connect with a broader sense of unity in nature and shared human experiences.
        \end{itemize}
        \vspace{5pt}
        \textbf{Note:}
        \begin{enumerate}[leftmargin=1.5em, nosep, label=\arabic*.]
            \item Based on the content provided, infer the likely motivation behind characters' behavior, if multiple motivations are possible, select all that apply.
            \item Select the best answer(s) from the options.
            \item Your output must be \textbf{ONLY the option letter(s) from the options provided choice :['A', 'B', 'C', 'D', 'E', 'F', 'G', 'H']} without explanation.
            \item ONLY output the option you choose, No explanation or justification is required.
        \end{enumerate}
        \vspace{5pt}
        THE ONLY output format is JSON String: \\
        \texttt{\{"answer": "option(s) that you choose"\}} \\
        \hline
    \end{tabular}
    \caption{The raw prompt input for the Maslow Practical task, containing 8 context-specific options generated based on Maslow's hierarchy.}
    \label{tab:maslow_practical_raw_prompt}
\end{table*}

\begin{table*}[h]
    \centering
    \small
    \renewcommand{\arraystretch}{1.2}
    \begin{tabular}{|p{0.95\linewidth}|}
        \hline
        \rowcolor{gray!15} \textbf{\large Maslow Definition Visual Reasoning Prompt} \\
        \hline
        \vspace{2pt}
        The following is a Visual Reasoning Multiple-Choice Question. Carefully read the story and view the images, fully immerse yourself in the role of the character described, and reason based on the information provided. Your answer should rely strictly on the given details. \\
        \vspace{5pt}
        \textbf{Story Context:} The tattooed woman with pink mohawk is excited and ready for her outdoor trip with her partner. The tattooed woman with pink mohawk is excitedly describing their outdoor trip plans. The tattooed woman with pink mohawk is riding in the vehicle, excited for the road trip. \\
        \vspace{5pt}
        \textbf{Question:} \\
        Based on the visual information and the story provided, which level(s) of Maslow's hierarchy of needs is/are most strongly expressed or fulfilled by the behavior of Tattooed woman with pink mohawk? \\
        \vspace{5pt}
        \textbf{Options:} \\
        A. Physiological needs : air, food, drink, shelter, warmth, sex, sleep, etc \\
        B. Safety needs : protection from elements, security, order, law, stability, freedom from fear. \\
        C. Love and belongingness needs : friendship, intimacy, trust, and acceptance, receiving and giving affection and love. Affiliating, being part of a group (family, friends, work). \\
        D. Esteem needs : which Maslow classified into two categories: (i) esteem for oneself (dignity, achievement, mastery, independence) and (ii) the need to be accepted and valued by others (e.g., status, prestige). \\
        E. Cognitive needs : knowledge and understanding, curiosity, exploration, need for meaning and predictability. \\
        F. Aesthetic needs : appreciation and search for beauty, balance, form, etc. \\
        G. Self-actualization needs : realizing personal potential, self-fulfillment, seeking personal growth, and peak experiences \\
        H. Transcendence needs : A person is motivated by values that transcend beyond the personal self. Beyond self-actualization, they represent the human desire to connect with a higher reality, purpose, or the universe. \\
        \vspace{5pt}
        \textbf{Note:}
        \begin{enumerate}[leftmargin=1.5em, nosep, label=\arabic*.]
            \item Based on the content provided, infer the likely motivation behind characters' behavior, if multiple motivations are possible, select all that apply.
            \item Select the best answer(s) from the options.
            \item Your output must be \textbf{ONLY the option letter(s) from the options provided choice :['A', 'B', 'C', 'D', 'E', 'F', 'G', 'H']} without explanation.
            \item ONLY output the option you choose, No explanation or justification is required.
        \end{enumerate}
        \vspace{5pt}
        THE ONLY output format is JSON String: \\
        \texttt{\{"answer": "option(s) that you choose"\}} \\
        \hline
    \end{tabular}
    \caption{The raw prompt input for the Maslow Definition task, including the story context, exact definitions, and formatting constraints.}
    \label{tab:maslow_definition_raw_prompt}
\end{table*}

\begin{table*}[h]
    \centering
    \small
    \renewcommand{\arraystretch}{1.2}
    \begin{tabular}{|p{0.95\linewidth}|}
        \hline
        \rowcolor{gray!15} \textbf{\large Reiss Practical Visual Reasoning Prompt} \\
        \hline
        \vspace{2pt}
        The following is a Visual Reasoning Multiple-Choice Question. Carefully read the story and view the images, fully immerse yourself in the role of the character described, and reason based on the information provided. Your answer should rely strictly on the given details. \\
        \vspace{5pt}
        \textbf{Story Context:} The tattooed woman with pink mohawk is excited and ready for her outdoor trip with her partner. The tattooed woman with pink mohawk is excitedly describing their outdoor trip plans. The tattooed woman with pink mohawk is riding in the vehicle, excited for the road trip. \\
        \vspace{5pt}
        \textbf{Question:} \\
        Based on the visual information and story provided, what is/ are mostly the motivation(s) of Tattooed woman with pink mohawk's behavior? \\
        \vspace{5pt}
        \textbf{Options:}
        \begin{itemize}[leftmargin=1.5em, nosep, label={}]
            \item \textbf{A.} Tattooed woman with pink mohawk wants to reach their outdoor destination to enjoy physical activities and movement.
            \item \textbf{B.} Tattooed woman with pink mohawk hopes to arrive at their trip spot to savor special foods and meals.
            \item \textbf{C.} Tattooed woman with pink mohawk feels driven to follow the planned itinerary for their road trip ritual.
            \item \textbf{D.} Tattooed woman with pink mohawk would like to complete the economical road trip to preserve their resources.
            \item \textbf{E.} Tattooed woman with pink mohawk hopes to escape daily stresses through the calming road trip journey.
            \item \textbf{F.} Tattooed woman with pink mohawk wants to share intimate moments with her partner during the road trip.
            \item \textbf{G.} Tattooed woman with pink mohawk hopes to create lasting family memories on this outdoor adventure.
            \item \textbf{H.} Tattooed woman with pink mohawk wants to gain approval from peers by embarking on this exciting trip.
            \item \textbf{I.} Tattooed woman with pink mohawk feels driven to enjoy companionship with her partner throughout the road trip.
            \item \textbf{J.} Tattooed woman with pink mohawk would like to assert her autonomy by choosing this spontaneous road trip path.
            \item \textbf{K.} Tattooed woman with pink mohawk hopes to outpace rival travelers by speeding ahead on the road trip.
            \item \textbf{L.} Tattooed woman with pink mohawk wants to uphold her commitment to promised trip traditions.
            \item \textbf{M.} Tattooed woman with pink mohawk feels driven to lead the navigation decisions during the road trip.
            \item \textbf{N.} Tattooed woman with pink mohawk hopes to elevate her adventurous image through this bold road trip.
            \item \textbf{O.} Tattooed woman with pink mohawk wants to discover new scenic views and places along the road trip.
            \item \textbf{P.} Tattooed woman with pink mohawk would like to connect with nature to support environmental appreciation.
        \end{itemize}
        \vspace{5pt}
        \textbf{Note:}
        \begin{enumerate}[leftmargin=1.5em, nosep, label=\arabic*.]
            \item Based on the content provided, infer the likely motivation behind characters' behavior, if multiple motivations are possible, select all that apply.
            \item Select the best answer(s) from the options.
            \item Your output must be \textbf{ONLY the option letter(s) from the options provided choice :['A', 'B', ..., 'P']} without explanation.
            \item ONLY output the option you choose, No explanation or justification is required.
        \end{enumerate}
        \vspace{5pt}
        THE ONLY output format is JSON String: \\
        \texttt{\{"answer": "option(s) that you choose"\}} \\
        \hline
    \end{tabular}
    \caption{The raw prompt input for the Reiss Practical task, containing 16 options generated based on Reiss's basic desires theory.}
    \label{tab:reiss_practical_raw_prompt}
\end{table*}

\begin{table*}[h]
    \centering
    \small
    \renewcommand{\arraystretch}{1.2}
    \begin{tabular}{|p{0.95\linewidth}|}
        \hline
        \rowcolor{gray!15} \textbf{\large Reiss Definition Visual Reasoning Prompt} \\
        \hline
        \vspace{2pt}
        The following is a Visual Reasoning Multiple-Choice Question. Carefully read the story and view the images, fully immerse yourself in the role of the character described, and reason based on the information provided. Your answer should rely strictly on the given details. \\
        \vspace{5pt}
        \textbf{Story Context:} The tattooed woman with pink mohawk is excited and ready for her outdoor trip with her partner. The tattooed woman with pink mohawk is excitedly describing their outdoor trip plans. The tattooed woman with pink mohawk is riding in the vehicle, excited for the road trip. \\
        \vspace{5pt}
        \textbf{Question:} \\
        Based on the visual information and the story provided, which of Reiss's 16 basic desires appear(s) to be the most strongly expressed or fulfilled by the behavior of Tattooed woman with pink mohawk? \\
        \vspace{5pt}
        \textbf{Options:}
        \begin{itemize}[leftmargin=1.5em, nosep, label={}]
            \item \textbf{A.} Physical Exercise : Drive for physical activity and movement
            \item \textbf{B.} Eating : Desire to eat
            \item \textbf{C.} Order : Desire to organize (including desire for ritual)
            \item \textbf{D.} Saving : Desire to collect, value of frugality
            \item \textbf{E.} Tranquility : Desire to avoid anxiety, fear
            \item \textbf{F.} Romance : Desire for sex (including courting)
            \item \textbf{G.} Family : Desire to raise own children
            \item \textbf{H.} Acceptance : Desire for approval
            \item \textbf{I.} Social Contact : Desire for peer companionship (desire to play)
            \item \textbf{J.} Independence : Desire to be autonomous
            \item \textbf{K.} Vengeance : Desire to get even (including desire to compete, to win)
            \item \textbf{L.} Honor : Desire to obey a traditional moral code
            \item \textbf{M.} Power : Desire to influence (including leadership; related to mastery)
            \item \textbf{N.} Status : Desire for social standing
            \item \textbf{O.} Curiosity : Desire for knowledge
            \item \textbf{P.} Idealism : Desire to improve society (including altruism, justice)
        \end{itemize}
        \vspace{5pt}
        \textbf{Note:}
        \begin{enumerate}[leftmargin=1.5em, nosep, label=\arabic*.]
            \item Based on the content provided, infer the likely motivation behind characters' behavior, if multiple motivation is possible, select all that apply.
            \item Select the best answer(s) from the options.
            \item Your output must be \textbf{ONLY the option letter(s) from the options provided choice :['A', 'B', 'C', 'D', 'E', 'F', 'G', 'H', 'I', 'J', 'K', 'L', 'M', 'N', 'O', 'P']} without explanation.
            \item ONLY output the option you choose, No explanation or justification is required.
        \end{enumerate}
        \vspace{5pt}
        THE ONLY output format is JSON String: \\
        \texttt{\{"answer": "option(s) that you choose"\}} \\
        \hline
    \end{tabular}
    \caption{The raw prompt input for the Reiss Definition task, asking the model to map behaviors directly to the 16 standard theoretical definitions.}
    \label{tab:reiss_definition_raw_prompt}
\end{table*}
\begin{table*}[h]
    \centering
    \small
    \renewcommand{\arraystretch}{1.3}
    \begin{tabularx}{\linewidth}{|l|X|l|}
        \hline
        \rowcolor{gray!15} \textbf{Mode} & \textbf{Introduction (Prompt Header)} & \textbf{Story Text Handling} \\
        \hline
        \textbf{Multimodal} & ``The following is a \textbf{Visual Reasoning} Multiple-Choice Question. Carefully \textbf{read the story and view the images}, fully immerse yourself in the role of the character described, and reason based on the information provided...'' & Includes full \texttt{Story Context}. \\
        \hline
        \textbf{Image-Only} & ``The following is a \textbf{Visual Reasoning} Multiple-Choice Question. Carefully \textbf{view the images}, fully immerse yourself in the role of the character described, and reason based on the information provided...'' & \textit{Hidden:} ``[Story context hidden for visual-only analysis]'' \\
        \hline
        \textbf{Text-Only} & ``The following is a \textbf{Reasoning} Multiple-Choice Question. Carefully \textbf{read the story}, fully immerse yourself in the role of the character described, and reason based on the information provided...'' & Includes full \texttt{Story Context}. \\
        \hline
    \end{tabularx}
    \caption{Prompt modifications for ablation studies. The introduction and context visibility are adjusted to restrict the model's input source for Image-Only and Text-Only evaluations.}
    \label{tab:modality_prompts}
\end{table*}

\end{document}